%% file: main.tex
\documentclass[10pt]{forithmus}

\usepackage{tikz}
\usetikzlibrary{arrows.meta,positioning}

\newif\ifnaturesubmission
\naturesubmissionfalse

\title{
Reliable and grounded reporting of any computed tomography scan with FORA
}
\shorttitle{FORA}

\author{Sezgin Er\textsuperscript{1,2,3,*},
Utku Turkbey\textsuperscript{1},
Alperen G\"ur\textsuperscript{1,4},
Omer Faruk Durugol\textsuperscript{5},
Suprosanna Shit\textsuperscript{2},
Murong Xu\textsuperscript{2},
Julia Wolleb\textsuperscript{2,4},
Jiayi Wang\textsuperscript{5},
Xiaoliang Wang\textsuperscript{5},
Hadrien Reynaud\textsuperscript{5},
Teodora Spasojevic\textsuperscript{5},
Dong Yang\textsuperscript{7},
Andriy Myronenko\textsuperscript{7},
Pengfei Guo\textsuperscript{7},
Yufan He\textsuperscript{7},
Can Zhao\textsuperscript{7},
Marc Edgar\textsuperscript{7,\ddag},
Andreas Maier\textsuperscript{5},
Michael Moor\textsuperscript{4},
Mehmet Kemal Ozdemir\textsuperscript{3},
Daguang Xu\textsuperscript{7},
Bernhard Kainz\textsuperscript{5,6,\dag},
Bjoern Menze\textsuperscript{2,*,\dag},
Ibrahim Ethem Hamamci\textsuperscript{1,2,3,*,\dag}}
\affiliation{%
\textsuperscript{1}Forithmus, San Francisco, CA, USA\\
\textsuperscript{2}University of Zurich, Zurich, Switzerland\\
\textsuperscript{3}Istanbul Medipol University, Istanbul, Turkiye\\
\textsuperscript{4}ETH Zurich, Zurich, Switzerland\\
\textsuperscript{5}Friedrich-Alexander-Universit\"at Erlangen-N\"urnberg (FAU), Erlangen, Germany\\
\textsuperscript{6}Imperial College London, London, UK\\
\textsuperscript{7}NVIDIA, Santa Clara, CA, USA}
\date{}
\correspondence{%
\textsuperscript{\*}Sezgin Er
(\href{mailto:sezgin.er@forithmus.com}{sezgin.er@forithmus.com}),
Bjoern Menze
(\href{mailto:bjoern.menze@uzh.ch}{bjoern.menze@uzh.ch})
and Ibrahim Ethem Hamamci
(\href{mailto:ibrahim.hamamci@forithmus.com}{ibrahim.hamamci@forithmus.com}).

\textsuperscript{\dag}These authors jointly supervised this work. \textsuperscript{\ddag}M.E. contributed to this work while employed at NVIDIA.
}

\abstracttext{Artificial intelligence (AI) systems for computed tomography (CT) remain largely restricted to specific anatomical regions, examination types or predefined abnormalities, limiting their ability to provide comprehensive and spatially grounded interpretations. Here we present FORA, a general-purpose CT model that recognizes and spatially localizes 1,028 hierarchically organized radiological concepts and generates reports across the 28 CT protocols represented in our cohort. A single model covers a broad range of abnormalities across anatomical regions, without separate models for individual organs, diseases or examination types. Trained on 304,962 routine CT examinations with supervision from their accompanying reports, FORA learns to localize abnormalities without manually annotated lesions. Its architecture combines multiple-instance learning with an interpretable concept bottleneck that connects spatial evidence, radiological findings and report generation. We first developed and benchmarked the architecture on chest radiographs, achieving state-of-the-art performance, before training a separate model on three-dimensional CT examinations. On 36,619 held-out CT examinations, FORA achieved a macro area under the receiver operating characteristic curve (AUROC) of 0.918 for radiological concept recognition, exceeding the strongest of 18 existing CT models by 0.21. FORA also ranked first on three public benchmarks, substantially outperforming Merlin on its own test set despite no training on data from that institution (AUROC 0.907 versus 0.826). On independent localization benchmarks for pulmonary emboli, renal masses and rib fractures, FORA's strongest spatial evidence fell within annotated lesions at rates up to 348 times greater than chance. FORA achieved more than threefold higher macro-averaged finding recall than the strongest existing report generator. Together, these results demonstrate that a single model can combine broad abnormality recognition, spatial localization and report generation across diverse CT examinations, using routine reports without lesion-level supervision.}

\usepackage{newfloat}
\usepackage{longtable}
\setcitestyle{numbers,super,comma,open={},close={}}

\makeatletter\def\NAT@sort{1}\def\NAT@cmprs{1}\makeatother
\DeclareFloatingEnvironment[fileext=loed,placement=p,name={Extended Data Figure}]{extendeddatafigure}
\crefname{extendeddatafigure}{Extended Data Figure}{Extended Data Figures}
\Crefname{extendeddatafigure}{Extended Data Figure}{Extended Data Figures}

\DeclareFloatingEnvironment[fileext=losf,placement=p,name={Supplementary Figure}]{supplementaryfigure}
\crefname{supplementaryfigure}{Supplementary Figure}{Supplementary Figures}
\Crefname{supplementaryfigure}{Supplementary Figure}{Supplementary Figures}

\begin{document}
\setlength{\emergencystretch}{1.5em}   
\maketitle

\section{Introduction}

Demand for medical imaging has grown steadily over the past two decades\cite{smithbindman2019}, and computed tomography (CT) use alone reached approximately 93 million examinations in the US in 2023\cite{smithbindman2025}. CT combines rapid volumetric acquisition and high spatial resolution with image intensities that represent the X-ray attenuation of tissue, linearly scaled to Hounsfield units\cite{mccollough2023}, making it essential for diagnosis across anatomical regions and clinical indications. Interpreting a single examination requires identifying abnormalities across thousands of images, involving multiple organs and disease processes, and the growing number of CT studies has substantially increased radiologist workload\cite{bruls2020}. Missed abnormalities remain an important source of diagnostic error\cite{bruno2015}.

Artificial intelligence (AI) systems can detect specific diseases\cite{ardila2019,cao2023} and generate reports from volumetric images\cite{ct2rep,radfm}. However, detecting a disease or producing a plausible report does not guarantee a thorough interpretation of an examination. Some important findings may be missed, and reported abnormalities may not have clear supporting evidence\cite{tumorvqa,deeptumorvqa}. A general-purpose CT model must therefore recognize a broad range of abnormalities across anatomical regions and examination types while linking individual findings to their locations in the scan.

\begin{figure}[!t]
  \centering
  \includegraphics[width=\linewidth]{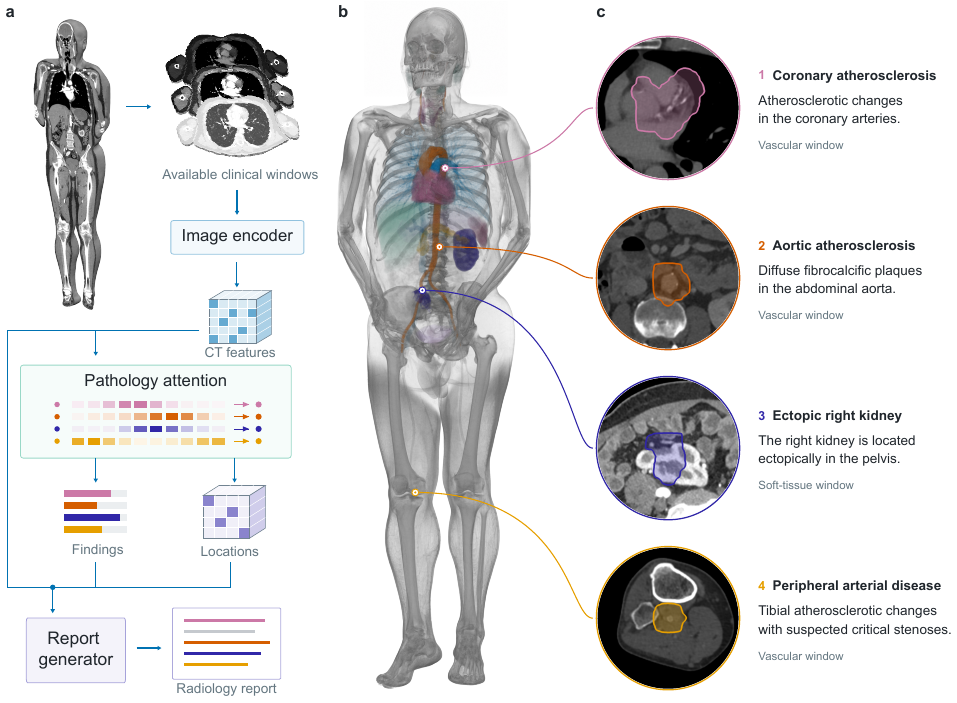}
  \caption{FORA links CT findings to spatial evidence and report text.
  \textbf{a}, Model overview. CT images are encoded across available clinical intensity windows (Hounsfield-unit ranges that display specific tissues, such as lung, soft tissue or bone), and concept-specific attention identifies findings and their locations to guide report generation.
  \textbf{b}, Example CT examination showing spatial evidence for multiple findings.
  \textbf{c}, Image regions supporting selected findings, paired with the corresponding generated report sentences.}
  \label{fig:overview}
\end{figure}
Research towards this goal has followed two largely separate directions. CT foundation models, including CT-CLIP\cite{ctclip}, Merlin\cite{merlin}, RADAR\cite{radar} and Pillar-0\cite{pillar0}, have extended recognition beyond individual diseases\cite{fvlm,colipri,sigvlp}, but covering the whole range of findings encountered across CT examination types remains challenging. In parallel, CT2Rep\cite{ct2rep}, CT-CHAT\cite{ctclip}, BTB3D\cite{btb3d}, RadFM\cite{radfm}, MedMPT\cite{medmpt} and NV-Reason-CT\cite{nvreasonct} have advanced report generation from volumetric images. However, coherent reports do not ensure that all relevant abnormalities have been recognized or that each statement is supported by image evidence. Grounded report generation links statements to image regions\cite{maira2,radgenome,unibiomed}, but methods requiring manually annotated bounding boxes or segmentation masks\cite{raddiagseg,rexgrounding} are difficult to scale to the diversity of findings in routine CT. Clinical reports have reduced the annotation burden for tumor segmentation\cite{scalingreports,rtsuper}, yet whether they can jointly supervise broad abnormality recognition, spatial localization and report generation remains unresolved.

Routine radiology reports usually describe findings for the whole examination, but they often do not specify exactly where these findings are located. Multiple-instance learning\cite{dietterich1997,ilse2018} can help identify which image regions support examination-level labels, so spatial evidence can be learned without needing lesion-level annotations. Radiological concepts also help connect images with language. Concept bottleneck models\cite{koh2020} use clear, human-understandable concepts to guide later predictions. Combining these concepts and their spatial evidence with visual features could guide report generation while preserving information beyond a predefined vocabulary.

Here we present FORA, a general-purpose CT model that recognizes and spatially localizes 1,028 hierarchically organized radiological concepts based on Radiopaedia\cite{radiopaedia} and uses these concepts alongside visual features to generate radiology reports (Fig.~\ref{fig:overview}). FORA combines a self-supervised three-dimensional visual encoder, attention-based multiple-instance learning and concept-guided report generation, with supervision from routine clinical reports and no manually annotated lesions. We first developed and evaluated the approach on 286,560 chest radiograph studies with corresponding reports, achieving state-of-the-art or comparable performance on established classification, localization and report-generation benchmarks (Extended Data Fig.~\ref{fig:ed-cxr}). We subsequently trained a separate model on volumetric CT using 304,962 examinations and evaluated it on 36,619 held-out examinations from a cohort of 341,581 examinations involving 238,469 patients. The CT cohort contains 28 acquisition protocol classes, grouped into 11 examination types and a mixed-examination category, yielding 12 categories for evaluation. A single CT model covers all 28 protocols without requiring separate models for individual organs or diseases.

FORA achieved a macro area under the receiver operating characteristic curve (AUROC) of 0.918 on held-out CT examinations, exceeding the strongest of 18 existing CT models by 0.21 (Fig.~\hyperref[fig:classification]{\ref*{fig:classification}a,b,e} and Extended Data Fig.~\ref{fig:ed-ct-types}). It also ranked first on three public benchmarks, substantially outperforming Merlin\cite{merlin} on its own test set without training on data from that institution (AUROC 0.907 versus 0.826; Fig.~\hyperref[fig:classification]{\ref*{fig:classification}c}). Despite receiving no lesion-level supervision, FORA localized spatial evidence within independently annotated pulmonary emboli, renal cysts and masses, rib fractures and chest findings at rates up to 348 times greater than chance (Fig.~\hyperref[fig:classification]{\ref*{fig:classification}f} and Extended Data Fig.~\ref{fig:ed-localization}). Its generated reports achieved more than threefold higher macro-averaged finding recall than the strongest of seven existing report generators (Fig.~\ref{fig:reportgen} and Extended Data Fig.~\ref{fig:ed-reportgen}). 
These results establish that routine radiology reports, which are produced for every clinical examination, are sufficient to train a single model that recognizes, localizes and reports abnormalities across 12 CT examination types, replacing lesion-level annotation and disease-specific models with supervision that already exists at scale.

\section{Results}
\label{sec:results}

\subsection{Data and Radiology concepts}

The model uses a concept bottleneck with 1,028 radiological concepts based on Radiopaedia\cite{radiopaedia}. For each examination, FORA predicts the probability of each concept and identifies supporting spatial evidence in the scan. These predictions guide report generation alongside visual features, linking recognition, localization and reporting through an explicit intermediate representation.

We collected structured labels from the routine radiology reports of 341,581 CT examinations from 238,469 patients. The scans use 28 scan protocols, which we grouped into 12 examination types for analysis. The cohort includes 136,269 chest, 82,525 head and neck, 74,556 abdominal and pelvic, 20,468 musculoskeletal, 16,459 angiographic and 11,304 multi-region or other examinations (Fig.~\hyperref[fig:dataset]{\ref*{fig:dataset}a}). We assembled a vocabulary of 1,459 concepts based on Radiopaedia\cite{radiopaedia}, covering abnormalities, anatomical variants, medical devices and postoperative states. The vocabulary is organized into a hierarchy of up to five levels, with broader concepts linked to more specific ones. Individual concepts can have multiple parents (Supplementary Fig.~\ref{fig:supplementary-1}).

To obtain structured labels, we developed a language-model-based pipeline that identifies concepts mentioned in each report, verifies them against quoted report text, and resolves conflicting labels (Supplementary Figs.~\hyperref[fig:supp-reportgen-labels]{\ref*{fig:supp-reportgen-labels}c} and~\hyperref[fig:supp-prompts-labels]{\ref*{fig:supp-prompts-labels}a,b}). Of the 1,459 concepts, 1,366 appeared in the reports, and 1,028 were included as FORA’s prediction targets, each with at least 20 positive training examinations. We refined the labelling pipeline using an audit of 2,403 labelled items from 215 reports, assessed by two independent language-model judges and an adjudicating model (Supplementary Fig.~\hyperref[fig:supp-reportgen-labels]{\ref*{fig:supp-reportgen-labels}c}). The extracted labels also allowed us to examine concept co-occurrence within and across examination types. The resulting network contains 587 concepts connected by 1,077 edges (Fig.~\hyperref[fig:dataset]{\ref*{fig:dataset}b}). The cohort includes scanners from three manufacturers, with variation in slice spacing, labels per examination, model-input extent, patient age, report length, tube voltage and radiation dose (Fig.~\hyperref[fig:dataset]{\ref*{fig:dataset}c-h,k,l}). We also examined the frequency of individual concepts and the proportions of asserted, uncertain and denied mentions (Fig.~\hyperref[fig:dataset]{\ref*{fig:dataset}i,j}).

FORA encodes all available reconstructions and intensity windows of each examination using a self-supervised three-dimensional vision transformer (Fig.~\hyperref[fig:overview]{\ref*{fig:overview}a} and Supplementary Fig.~\ref{fig:supp-architecture}). An attention-based multiple-instance learning head predicts each concept from image tokens across the examination and assigns attention to the regions contributing to its prediction. These concept-specific attention maps provide spatial evidence without requiring lesion-level annotations during training. Model weights, calibration parameters and decision thresholds were determined using training data only (Supplementary Fig.~\ref{fig:supp-evaluation}).

\subsection{Recognition performance exceeds existing CT models in every examination type}

On held-out examinations, FORA achieved a macro area under the receiver operating characteristic curve (AUROC) of 0.918 (95\% CI, 0.916-0.919), compared with 0.705 for CoLiPri\cite{colipri}, the strongest of 18 comparison models. FORA also achieved higher area under the precision-recall curve (AUPRC; 0.393 versus 0.146), balanced accuracy (0.837 versus 0.630) and CRG\cite{crg} (0.705 versus 0.431; all $P<0.001$; Fig.~\hyperref[fig:classification]{\ref*{fig:classification}a}). The improvement extended across findings and examination types. FORA achieved higher average precision in 2,040 of 2,132 concept-examination-type pairs (95.7\%; Fig.~\hyperref[fig:classification]{\ref*{fig:classification}b}). Its median AUROC advantage over the strongest alternative was positive in every examination type, ranging from 0.07 for chest CT to 0.21 for sinus and orbit CT (Fig.~\hyperref[fig:classification]{\ref*{fig:classification}e}). FORA also ranked first across all 12 examination types on each of the four metrics (Fig.~\hyperref[fig:classification]{\ref*{fig:classification}d} and Extended Data Fig.~\ref{fig:ed-ct-types}).

\input{fig2-classification-localization-float}

We next examined performance across concepts with different frequencies in the training data. Average precision exceeded concept prevalence for all 1,019 concepts with at least one positive held-out patient, by a median factor of 98. Absolute average precision increased with the number of positive patients, from 0.08 for concepts with fewer than 20 positive patients to 0.75 for those with at least 1,000 (Supplementary Fig.~\hyperref[fig:fora-supp2-classifier-reliability]{\ref*{fig:fora-supp2-classifier-reliability}a,b}). AUROC showed a weak negative association with the number of training positives (Spearman's $\rho=-0.20$ for CT and $-0.14$ for chest radiographs). Median AUROC was 0.944 within individual examination types and ranged from 0.94 to 0.99 across organ systems (Supplementary Fig.~\hyperref[fig:fora-supp2-classifier-reliability]{\ref*{fig:fora-supp2-classifier-reliability}c-g}). Mean predicted and observed concept prevalences were 1.09\% and 1.10\% for CT, respectively, and 0.91\% and 0.92\% for chest radiographs. The median expected calibration error was 0.0003 per concept (Supplementary Fig.~\ref{fig:supp-calibration}).

FORA also achieved the highest AUROC on three public benchmarks, including two from institutions outside its training hospital network (Fig.~\hyperref[fig:classification]{\ref*{fig:classification}c}). On the Merlin test set\cite{merlin}, which originates from an institution that also contributed to Merlin's training data, FORA achieved an AUROC of 0.907 compared with 0.826 for Merlin. FORA achieved higher AUROC and average precision for 27 of the 30 evaluated concepts (Supplementary Fig.~\hyperref[fig:fora-supp3-transfer-training]{\ref*{fig:fora-supp3-transfer-training}a-d}). For the 21 concepts also covered by RADAR\cite{radar}, which combines organ segmentation with organ-specific finding recognition, FORA achieved an AUROC of 0.916 compared with 0.885 for RADAR when evaluated using its published protocol (Supplementary Fig.~\hyperref[fig:fora-supp3-transfer-training]{\ref*{fig:fora-supp3-transfer-training}e}). FORA also achieved the highest AUROC on RAD-ChestCT\cite{radchestct} (0.816 versus 0.796) and CT-RATE\cite{ctclip} (0.891 versus 0.871 for NV-Reason-CT\cite{nvreasonct}). RAD-ChestCT originates from another institution, whereas CT-RATE shares FORA's hospital network; patients in the CT-RATE validation set were excluded from FORA's training data. On two additional single-finding datasets, FORA achieved an AUROC of 0.947 for rib fractures in RibFrac\cite{ribfrac}, compared with 0.818 for the strongest alternative, and an AUROC of 0.794 for renal cysts in KiTS23\cite{kits21}, exceeding all comparison models except RADAR (0.850; Supplementary Fig.~\hyperref[fig:fora-supp3-transfer-training]{\ref*{fig:fora-supp3-transfer-training}e}).

We further assessed performance across patient and acquisition subgroups. Macro AUROC differed by 0.002 between sexes, 0.02 across scanner manufacturers and 0.005 across tube voltages. Across age groups, AUROC ranged from 0.873 to 0.925 for CT and from 0.871 to 0.928 for chest radiographs (Supplementary Fig.~\ref{fig:supp-subgroups}).

\subsection{Report supervision yields spatially accurate evidence}

FORA produces a spatial evidence map for each of its 1,028 radiological concepts. We evaluated these maps using independent lesion annotations from four datasets covering pulmonary emboli, renal masses, renal cysts, rib fractures and chest findings. None of these annotations was used during training or to optimize the localization method. The centre of the strongest evidence focus, the top focus, fell within the annotated lesion substantially more often than expected by chance across all four datasets. The hit rate was 69.7\% for pulmonary emboli in FUMPE\cite{fumpe}, 72.8\% for renal masses and 66.1\% for renal cysts in KiTS23, and 58.3\% for rib fractures in RibFrac, corresponding to 348, 10, 41 and 188 times the probability of chance placement within the same anatomical region, respectively (Fig.~\hyperref[fig:classification]{\ref*{fig:classification}f} and Extended Data Fig.~\hyperref[fig:ed-localization]{\ref*{fig:ed-localization}a}). On ReXGroundingCT\cite{rexgrounding}, which includes a broader range of chest findings, the top focus fell within the annotation in 27.6\% of instances, 33 times the chance expectation. When considering the region containing 80\% of the evidence rather than the top focus alone, overlap with the annotation was observed in 80.8-97.0\% of instances across the four datasets (Extended Data Fig.~\hyperref[fig:ed-localization]{\ref*{fig:ed-localization}a}).

We compared FORA with RADAR\cite{radar}, which localizes 146 findings within 36 segmented organs, and GLINT-CT\cite{glint}, which learns to segment findings described in free text from CT scans and reports without segmentation masks. All three models were evaluated using the same localization metrics across the four datasets, with lung nodules and pleural effusions additionally assessed separately. Targets unsupported by a model were assigned a score of zero (Extended Data Fig.~\hyperref[fig:ed-localization]{\ref*{fig:ed-localization}d}). FORA achieved higher top-focus hit rates than RADAR across all four datasets. For renal cysts, renal masses and rib fractures, the respective hit rates were 66.1\%, 72.8\% and 58.3\% for FORA, compared with 4.8\%, 36.4\% and 0\% for RADAR. At equal evidence-region sizes, FORA also achieved higher annotation-overlap rates in all settings except renal masses, where performance was comparable (95.7\% versus 96.1\%). Compared with GLINT-CT, evaluated using the threshold that maximized Dice score on each dataset as in its original study, FORA achieved higher top-focus hit rates for pulmonary emboli (69.7\% versus 0\%), renal cysts (66.1\% versus 29.0\%), renal masses (72.8\% versus 53.0\%) and rib fractures (58.3\% versus 40.0\%). Performance was similar across ReXGroundingCT findings (27.6\% versus 27.9\%). 
At equal evidence-region sizes, 
FORA had higher annotation-overlap rates than GLINT-CT for pulmonary emboli, renal cysts, renal masses, rib fractures, and pleural effusions, and similar results for lung nodules and overall ReXGroundingCT.

We further examined localization across different findings and lesion sizes. On ReXGroundingCT, top-focus hit rates were above chance in ten out of eleven categories. For mosaic attenuation, a diffuse finding without a clear focal centre, the evidence region matched the annotation in 57.1\% of instances (Extended Data Fig.~\hyperref[fig:ed-localization]{\ref*{fig:ed-localization}b}). Localization improved with lesion size. Top-focus hit rates increased from 10.6\% for annotations smaller than 0.5\,mL to 42.2\% for those of at least 50\,mL, while evidence-region overlap increased from 63.0\% to 93.5\% (Extended Data Fig.~\hyperref[fig:ed-localization]{\ref*{fig:ed-localization}c}). To illustrate localization beyond the independently annotated datasets, Supplementary Fig.~\ref{fig:supp-gallery} presents evidence maps for 12 additional findings outside the chest, including intracerebral haemorrhage, sinus mucosal thickening, hydronephrosis, abdominal aortic aneurysm and vertebral compression fracture, using the highest- and median-probability true-positive examinations for each concept.

\subsection{Concept-guided generation improves CT report completeness}

Our model generates radiology reports using predicted radiological concepts and their spatial evidence alongside visual features from the CT examination. The language model receives the recognized concepts with their predicted locations, sizes and supporting evidence, together with visual tokens, and is trained using supervised fine-tuning followed by direct preference optimization (DPO)\cite{dpo} (Fig.~\hyperref[fig:overview]{\ref*{fig:overview}b,c} and Supplementary Figs.~\hyperref[fig:supp-reportgen-labels]{\ref*{fig:supp-reportgen-labels}a,b} and~\ref{fig:supp-prompts-writers}). This provides the language model with explicit information about individual findings while retaining access to the underlying image representations.

We compared FORA with seven CT report generators, each with the prompt, decoding settings and token limit released by its developers.
Reports were scored using a common language-model-based finding extractor and a separate LLM judge (Fig.~\ref{fig:reportgen} and Supplementary Figs.~\hyperref[fig:supp-prompts-labels]{\ref*{fig:supp-prompts-labels}a,c} and~\ref{fig:supp-prompts-judge}). Compared with NV-Reason-CT, the strongest alternative, FORA achieved higher macro recall (0.215 versus 0.066), precision (0.285 versus 0.132), balanced accuracy (0.606 versus 0.532) and CRG (0.467 versus 0.387). Its judge score was also higher (0.570; 95\% CI, 0.566-0.573; versus 0.405; Fig.~\hyperref[fig:reportgen]{\ref*{fig:reportgen}b}). In direct pairwise comparisons, the judge preferred FORA's report for 73.8\% of internal examinations and NV-Reason-CT's for 23.9\%. FORA's preference rate ranged from 59.5\% for chest CT to 97.2\% for head and neck CT (Extended Data Fig.~\hyperref[fig:ed-reportgen]{\ref*{fig:ed-reportgen}c}). In the example chest CT shown in Fig.~\hyperref[fig:reportgen]{\ref*{fig:reportgen}a}, FORA recovered seven of eight reference findings, whereas no other generator recovered more than three.

\input{fig3-report-generation-float}

The improvement extended across body regions, acquisition protocols and report complexity. FORA outperformed all seven generators on all five evaluation measures in each of five body regions (all $P<0.01$; Extended Data Fig.~\hyperref[fig:ed-reportgen]{\ref*{fig:ed-reportgen}a}). Across the 12 most frequent report protocols, FORA achieved the highest CRG and judge scores, as well as the highest or equal balanced accuracy and recall (Fig.~\hyperref[fig:reportgen]{\ref*{fig:reportgen}c}). It also achieved the highest CRG and judge scores across all 17 protocols with at least 200 examinations and the pooled remaining protocols (Extended Data Fig.~\hyperref[fig:ed-reportgen]{\ref*{fig:ed-reportgen}b}). FORA ranked first across all levels of report complexity, and the number of findings in its generated reports increased with the number documented in the reference reports (Fig.~\hyperref[fig:reportgen]{\ref*{fig:reportgen}d}).

We also evaluated report generation on external benchmarks. On the Merlin test set, FORA outperformed all comparison generators, including Merlin's report generator trained at the institution from which the test set originated. FORA achieved a judge score of 0.325 compared with 0.260 for the strongest alternative, and approximately twice its recall (0.101 versus 0.050). On CT-RATE, which was used to train NV-Reason-CT, FORA achieved higher recall (0.277 versus 0.233) and balanced accuracy (0.623 versus 0.603), while matching NV-Reason-CT on precision and CRG. Judge scores were also comparable (0.675 versus 0.679; Fig.~\hyperref[fig:reportgen]{\ref*{fig:reportgen}e}). On internal validation, 32.3\% of FORA's reports received a perfect judge score, increasing to 50.6\% for head and neck CT (Extended Data Fig.~\hyperref[fig:ed-reportgen]{\ref*{fig:ed-reportgen}d}). A breakdown of the judge assessments is provided in Extended Data Fig.~\hyperref[fig:ed-reportgen]{\ref*{fig:ed-reportgen}e}. Supplementary Fig.~\ref{fig:supp-report-examples} presents four additional examinations with reports from all eight generators.

To assess the contribution of preference optimization, we compared the supervised writer before and after DPO using identical finding plans and evaluation procedures (Supplementary Fig.~\ref{fig:supp-dpo}). On 500 CT validation examinations, the supervised writer omitted 72.1\% of the reference findings included in its input plan, compared with 40.8\% after DPO. Micro-averaged recall increased from 0.255 to 0.531 and F1 from 0.385 to 0.592 (Supplementary Fig.~\hyperref[fig:supp-dpo]{\ref*{fig:supp-dpo}a}). The mean number of findings per generated report increased from 2.8 to 7.1, compared with 9.6 in the reference reports (Supplementary Fig.~\hyperref[fig:supp-dpo]{\ref*{fig:supp-dpo}b}), and omissions decreased across every body region (Supplementary Fig.~\hyperref[fig:supp-dpo]{\ref*{fig:supp-dpo}c}). On 28,965 chest radiograph validation examinations, the same training procedure increased recall from 0.262 to 0.417, balanced accuracy from 0.547 to 0.592 and CRG from 0.403 to 0.459 (Supplementary Fig.~\hyperref[fig:supp-dpo]{\ref*{fig:supp-dpo}d,e}).

\subsection{Development and benchmarking on chest radiographs}

We first developed and systematically evaluated FORA's architecture on chest radiographs before adapting it to volumetric CT. Chest radiographs provided established public benchmarks for classification, localization and report generation, allowing us to compare architectural choices and evaluate all three tasks against released models (Extended Data Fig.~\ref{fig:ed-cxr}). The radiograph model was trained on 257,595 studies using labels extracted from routine reports and evaluated on 28,965 internal validation studies and five public test sets. These experiments established the performance of the approach on two-dimensional images before its extension to CT, rather than evaluating a CT-trained model on radiographs.

For classification, FORA matched or exceeded the strongest of seven released models across all six evaluation sets (Extended Data Fig.~\hyperref[fig:ed-cxr]{\ref*{fig:ed-cxr}b}). On internal validation, FORA achieved a macro AUROC of 0.954 (95\% CI, 0.949-0.960) across the 128 of its 129 radiological concepts with validation positives, compared with 0.866 for RadZero\cite{radzero}. It achieved higher average precision for 108 of the 128 concepts and ranked first on AUROC, AUPRC, balanced accuracy and CRG across all ten concept categories (Extended Data Fig.~\hyperref[fig:ed-cxr]{\ref*{fig:ed-cxr}c,d}). On public benchmarks, FORA achieved AUROCs of 0.877 on RSNA Pneumonia\cite{rsna}, 0.865 on Open-I\cite{openi} and 0.813 on ChestX-Det10\cite{chestxdet10}, compared with 0.864, 0.847 and 0.796 for the strongest alternatives, respectively. Performance was comparable to CARZero\cite{carzero} on ChestX-ray14\cite{chestxray14} (0.811 versus 0.811) and PadChest-GR\cite{padchestgr} (0.863 versus 0.862).

FORA is able to localize findings without using bounding-box supervision. Across four test sets with radiologist-annotated boxes, the peak of the strongest evidence focus fell within the annotation for 66.9-89.2\% of finding instances, corresponding to five to ten times the chance expectation. FORA achieved higher localization hit rates than all comparison models producing spatial maps, with statistically significant differences in 22 of 24 comparisons (Extended Data Fig.~\hyperref[fig:ed-cxr]{\ref*{fig:ed-cxr}e,f}). In report generation, FORA recovered more reference findings than six comparison generators (macro recall 0.321 versus 0.213) and had fewer clinically significant errors per examination (0.47 versus 0.57; Extended Data Fig.~\hyperref[fig:ed-cxr]{\ref*{fig:ed-cxr}g}). Recall remained between 0.34 and 0.45 across groups stratified by the number of reference findings, compared with a maximum of 0.27 for the other generators (Extended Data Fig.~\hyperref[fig:ed-cxr]{\ref*{fig:ed-cxr}h}). In the example shown in Extended Data Fig.~\hyperref[fig:ed-cxr]{\ref*{fig:ed-cxr}a}, FORA recovered four of six reference findings, whereas no other generator recovered more than one.

\subsection{Learned representations encode tissue measurements and stable patient-specific profiles}

We examined how FORA's predicted radiological concepts characterize examinations across the cohort. Clustering the 1,028-dimensional concept profiles identified 16 groups with distinct anatomical distributions and patterns of reported findings (Extended Data Fig.~\ref{fig:fora-ed2-global-atlas}). These included groups enriched for turbinate hypertrophy, brain atrophy, coronary stenosis, urinary calculi and fractures. To examine variation within anatomical regions, we repeated the analysis separately for head, chest, abdomen, neck, cardiac and spine CT. Each region contained six groups with different combinations of findings, including pulmonary nodules and pleural effusions in chest CT, urinary calculi and ovarian cysts in abdominal CT, and spinal stenosis and compression fractures in spine CT (Extended Data Fig.~\ref{fig:fora-ed3-regional-atlas}).

\begin{figure}[!t]
  \centering
  \includegraphics[width=\linewidth]{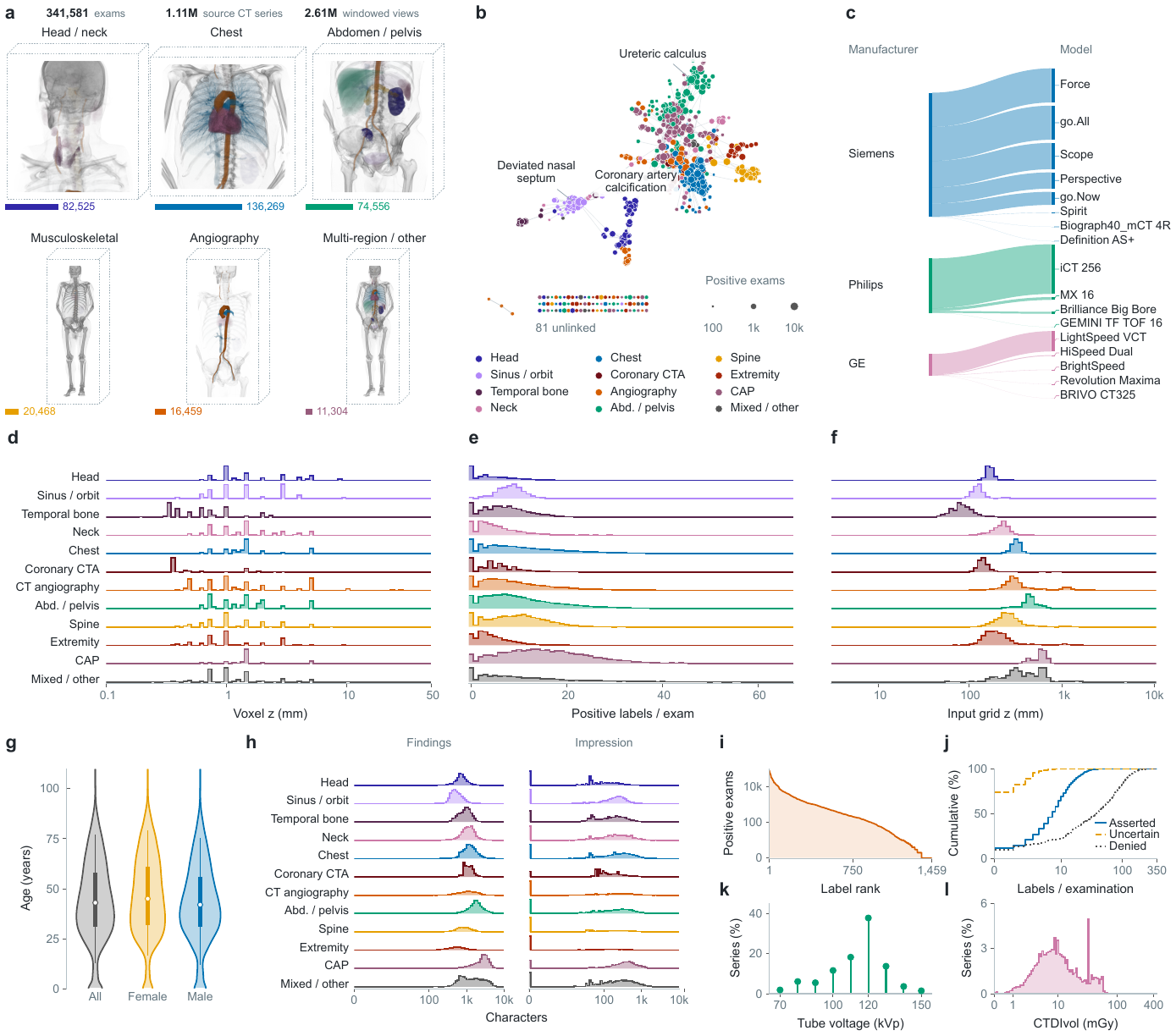}
  \caption{FORA cohort and report-derived supervision across 341,581 CT examinations from 238,469 patients.
  \textbf{a}, Examination counts by CT type. \textbf{b}, Finding co-occurrence network. \textbf{c}, Scanner manufacturers and models.
  \textbf{d-f}, Slice spacing, positive labels per examination and model-input extent by CT type. \textbf{g}, Patient age. \textbf{h}, Report length by CT type.
  \textbf{i}, Label frequencies across the 1,459 concepts in the finding hierarchy. \textbf{j}, Asserted, uncertain and denied labels per report.
  \textbf{k,l}, Tube voltage and volume CT dose index (CTDIvol).}
  \label{fig:dataset}
\end{figure}

We next examined whether FORA's visual representations retained quantitative anatomical information not explicitly included in its concept vocabulary. Using 12,700 examinations, we fitted out-of-fold linear readouts to predict liver attenuation, L3 muscle attenuation, aortic calcification and L1 vertebral attenuation from features of the self-supervised image encoder. Predicted values correlated with the corresponding measurements, with Pearson correlation coefficients of 0.990, 0.972, 0.861 and 0.819, respectively (Supplementary Fig.~\hyperref[fig:fora-supp4-quantitative-information]{\ref*{fig:fora-supp4-quantitative-information}a-d}). The readouts also tracked within-patient changes between examinations, with Spearman correlations ranging from 0.51 to 0.94 (Supplementary Fig.~\hyperref[fig:fora-supp4-quantitative-information]{\ref*{fig:fora-supp4-quantitative-information}e-h,j}). Readouts based on features from the relevant anatomical region had lower prediction errors than those using age, sex and tube voltage or features from another organ (Supplementary Fig.~\hyperref[fig:fora-supp4-quantitative-information]{\ref*{fig:fora-supp4-quantitative-information}i}). Bland-Altman analyses of agreement between predicted and measured values are shown in Supplementary Fig.~\hyperref[fig:fora-supp4-quantitative-information]{\ref*{fig:fora-supp4-quantitative-information}k-n}.

We also assessed whether FORA's concept predictions remained consistent across repeated examinations of the same patient. The cohort included 45,669 patients scanned on at least two dates, with 56,395 intervals between examinations of the same type (Extended Data Fig.~\ref{fig:extended-data-1}). Among 3,009 matched pairs, concept profiles from repeated examinations of the same patient had a mean cosine similarity of 0.879, compared with 0.754 for matched examinations from different patients, a difference of 0.126 (95\% CI, 0.122-0.130; Supplementary Fig.~\hyperref[fig:fora-supp5-repeat-consistency]{\ref*{fig:fora-supp5-repeat-consistency}a}). The difference was observed in six of seven CT categories and persisted when imaging site, contrast administration or scanner manufacturer differed between examinations (0.130 under unchanged conditions versus 0.118 when conditions changed; Supplementary Fig.~\hyperref[fig:fora-supp5-repeat-consistency]{\ref*{fig:fora-supp5-repeat-consistency}b,e}). It was 0.155 for examinations performed within 30 days and remained 0.098 for those separated by more than one year. Repeated examinations belonged to the same concept-profile group in 66-74\% of pairs, compared with 50-59\% of matched examinations from different patients (Supplementary Fig.~\hyperref[fig:fora-supp5-repeat-consistency]{\ref*{fig:fora-supp5-repeat-consistency}c,d,f,g}). Thus, FORA's visual representations retain quantitative tissue information, while its concept predictions capture anatomical and patient-specific patterns that remain relatively consistent across examinations.

\section{Discussion}
\label{sec:discussion}

CT models have progressed from single-disease detectors to foundation models trained on large collections of images and reports, including CT-CLIP\cite{ctclip}, Merlin\cite{merlin} and Pillar-0\cite{pillar0}. These models have broadened the range of recognizable abnormalities, but recognition, spatial localization and report generation have largely been studied as separate tasks. Clinical CT interpretation requires all three across examinations that differ substantially in anatomy, acquisition and pathology. FORA brings these capabilities together in a single model, recognizing 1,028 radiological concepts across 28 scan protocols, associating them with spatial evidence and using the predicted concepts alongside visual features to generate reports.

FORA was trained on 304,962 examinations using routine clinical reports without lesion-level annotations (Fig.~\hyperref[fig:training]{\ref*{fig:training}}). It performed better than 18 comparison models in 95.7\% of the evaluated concept-examination-type pairs, and these improvements also appeared in external benchmarks, including the Merlin test set. Its spatial evidence maps identified independently annotated abnormalities at rates well above chance in five evaluation settings, and its generated reports recovered approximately three times as many documented findings as the strongest comparison generator. The approach was first developed on chest radiographs, where it also matched or set the standard in classification, localization and report-generation benchmarks. Thus, our results indicate that broad concept recognition, weakly supervised localization and report generation can be combined across different imaging modalities without requiring separate models for individual diseases.

Several parts of our approach contribute to these capabilities. First, FORA uses explicit supervision for a large set of radiological concepts taken from clinical reports. Unlike contrastive image-text models, which typically infer individual findings through similarities between image and text representations, FORA directly predicts each concept from image features. This allows the model to represent findings independently of the wording used to describe them at inference time, although the quality and frequency of their documentation still affect training. Furthermore, attention-based multiple-instance learning associates examination-level concept labels with image regions. Anatomical routing focuses the search for supporting evidence on relevant regions but still keeps information from the whole examination. This lets the system learn spatial evidence maps without needing lesion-level supervision. FORA uses these clear concept predictions in report generation, along with visual tokens. The concept pathway gives a structured way to represent recognized abnormalities and their supporting evidence. The visual pathway lets the language model use information outside the set vocabulary. Preference optimization also made reports more complete by reducing missed findings from the generation plan.

The combination of broad recognition and spatial evidence could support several aspects of clinical CT interpretation. Predicting a large vocabulary of abnormalities may help identify incidental findings outside the primary clinical question, including abnormalities that might otherwise receive less attention during a focused examination. Linking concept predictions to image regions could also make AI-generated reports easier to review, allowing radiologists to inspect the evidence associated with individual findings. 
These predictions could help prioritize worklists for a wider range of abnormalities than systems focused only on specific emergencies. However, using them for triage would need future studies to check sensitivity, calibration, and clinical impact. FORA's visual outputs also keep quantitative details about tissue attenuation and calcification, which could be useful for assessing existing CT scans opportunistically\cite{pickhardt2020}. Using a single model for different exam types might also make it unnecessary to manage separate disease-specific systems, as long as its performance is proven for the intended clinical use.

The representation analyses provide further insight into what FORA learns from routine examinations. Predicted concept profiles separated examinations into groups characterized by different anatomical regions and combinations of findings, while quantitative tissue measurements could be recovered from the visual encoder using linear readouts. Repeated examinations from the same patient also had more similar concept profiles than matched examinations from different patients, including when acquisition conditions changed or scans were separated by more than a year. These observations suggest that FORA retains both continuous anatomical information and relatively stable patient-specific imaging characteristics. They do not establish that the model can assess disease progression, but they provide a basis for extending it to longitudinal interpretation. Incorporating prior examinations could allow future models to distinguish new, persistent and resolving abnormalities rather than interpreting each study independently.

Several limitations remain. FORA was trained on examinations from one hospital network, with reports translated into English. Although performance was evaluated on external datasets, differences in patient populations, imaging protocols and reporting practices may affect generalization to other healthcare systems\cite{benchx}. The concept labels were extracted from routine reports and therefore inherit their omissions, uncertainty and institutional reporting conventions. In particular, an unmentioned finding cannot always be assumed to be absent. The localization experiments used independent annotations across several anatomical regions, but quantitative validation was limited to concepts with suitable annotations. 
Accurate classification does not guarantee accurate localization\cite{saporta2022}, and evidence maps for concepts without suitable annotations have not been systematically validated. FORA's evidence decomposes the study logit exactly into regional contributions\cite{javed2022}, but these contributions are distinct from causal dependence, which would require interventional tests such as occluding the highlighted region\cite{jain2019attention,wiegreffe2019attention}. 
For report generation, we mainly used an automated finding extractor and language-model judges. These tools help compare models consistently, but they cannot replace independent clinical review.

Clinical validation is needed to determine whether these improvements translate into more accurate and efficient radiological interpretation and downstream patient benefits. Radiologist assessment of finding-level accuracy, report completeness, clinically significant errors and spatial evidence will be important, as will reader studies measuring the effect of FORA assistance on interpretation time and diagnostic performance\cite{tanno2025}. Future work could also extend the concept vocabulary and investigate the approach in other volumetric modalities, including MRI, where routine reports provide examination-level supervision but detailed spatial annotations remain limited. FORA shows that routine radiology reports can help recognize many radiological concepts and their locations, and these insights can support report generation along with visual features. By bringing these abilities together across different CT scans, this method lays the groundwork for more general-purpose systems in radiological interpretation.

\section{Methods}
\label{sec:methods}

\subsection{Cohort and data preparation}

We retrospectively collected 341,581 CT examinations from 238,469 patients at one hospital network. Each examination comprised a radiology report and the images described in that report. Examinations were assigned to 28 protocol classes. 27 CT protocols and dental cone-beam CT, which accounted for 65 examinations. These classes were grouped into 11 examination types and a mixed category for examinations spanning multiple types (Fig.~\ref{fig:dataset}). We split the cohort by patient into a training set of 304,962 examinations from 213,106 patients and a held-out set of 36,619 examinations from 25,363 patients, with no patient overlap. The training set excluded all CT-RATE\cite{ctclip} validation studies, including the 414 ReXGroundingCT\cite{rexgrounding} studies used for evaluation (Supplementary Fig.~\hyperref[fig:supp-evaluation]{\ref*{fig:supp-evaluation}a}).

Each usable series was resampled to a voxel spacing of 1.5 $\times$ 0.5 $\times$ 0.5 mm, cropped to the body and represented by multiple intensity views. These included the full range from $-1{,}000$ to $1{,}000$ HU, each distinct window stored in the DICOM header and, when a lung window with a center below $-250$ HU was present, a 10 mm lung maximum-intensity projection. All views from an examination formed its study bag. The training set contained 856,181 series and 2,332,070 views (Supplementary Fig.~\ref{fig:supplementary-1}). Reports were anonymized, translated from Turkish into English using Gemma-4-31B-it\cite{gemma4} and structured into sections. Findings and Impression sections were used for labeling.

\subsection{Report labels and finding hierarchy}

Each report was labeled against a closed vocabulary of 1,459 radiological concepts, which covers abnormalities, anatomical variants, devices and postoperative states and which we refer to as findings. Its 1,286 specific concepts were extracted from Radiopaedia\cite{radiopaedia} and extended with findings that radiologists reported in the corpus, such as devices and postoperative states. We arranged them in a hierarchy of up to five levels by adding 173 umbrella concepts, including the classes of the CT-RATE\cite{ctclip} and Merlin\cite{merlin} benchmarks and groups by organ and disease process, and by linking each concept to the more general concepts it implies. A concept can have several parents, and 150 of the 1,366 concepts that occur in the corpus have more than one (Supplementary Fig.~\ref{fig:supplementary-1}). Gemma-4-31B-it labeled the Findings and Impression sections in three passes. The first extracted candidate findings from the vocabulary, the second verified each against a verbatim quote from the report, and the third applied the verification and re-checked commonly missed findings. Each label received the state asserted, uncertain or denied (Supplementary Figs.~\hyperref[fig:supp-reportgen-labels]{\ref*{fig:supp-reportgen-labels}c} and~\hyperref[fig:supp-prompts-labels]{\ref*{fig:supp-prompts-labels}a,b}). Positive labels were combined with those of an earlier extraction with the same rules, and free-text findings whose name matched a vocabulary entry were added as asserted.

Further passes with the same model refined the labels. A recall pass added 197,769 asserted labels in 145,484 reports, each confirmed by a separate verification call with a verbatim quote. A malignancy check flagged cancer labels that referred only to history (10,804) or were unsupported (952), an adjudication pass settled conflicts between positive and denied labels, and a mapping pass assigned free-text findings to the vocabulary. Remaining conflicts were resolved in favour of the positive label. Asserted and uncertain labels were then propagated to all ancestors in the hierarchy and denials to all descendants. A finding counted as positive when it was asserted or uncertain, and findings that a report did not mention were treated as negative. We audited an earlier version of the extraction with two independent language-model judges and an adjudicating model, covering 2,403 labelled items in 215 reports. The audit found recurring errors, which we used to create prompt rules and guide the recall pass.

Findings with at least 200 positive and 200 negative training examinations (749) received their own classifier head. Findings with 20 to 199 positives (279) were attached to their nearest ancestor with its own head, initialized from it and regularized towards it during training. This gave 1,028 targets. Eighty-six rare findings without such an ancestor and two findings that are not visible on CT were not modeled.

\subsection{Image encoder}

\begin{figure}[!t]
  \centering
  \includegraphics[width=\linewidth]{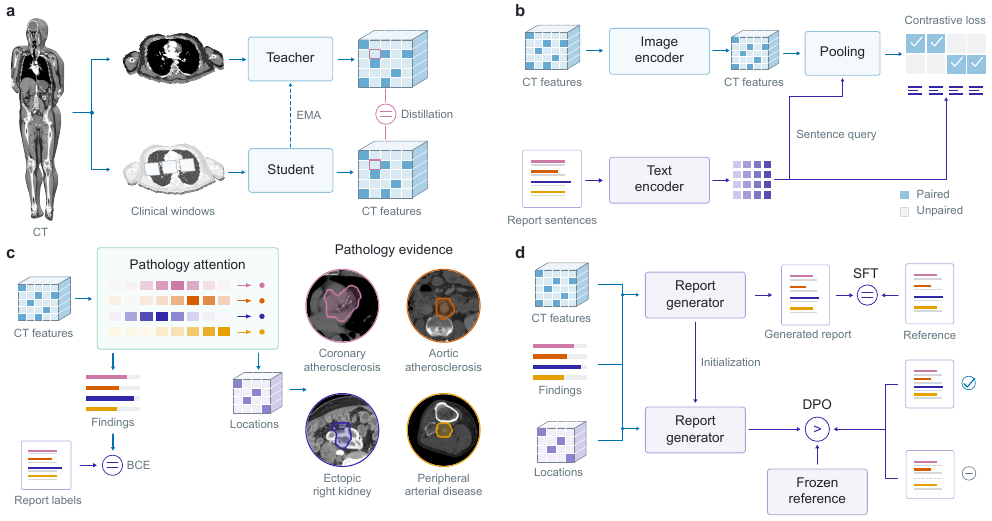}
  \caption{FORA learns finding recognition, spatial localization and report generation from routine CT examinations and reports.
  \textbf{a}, Self-supervised teacher--student distillation across clinical intensity windows.
  \textbf{b}, Contrastive alignment of CT features with report sentences.
  \textbf{c}, Concept-specific attention learns finding recognition and spatial evidence from report-derived labels.
  \textbf{d}, Concept-guided report generation trained through supervised fine-tuning (SFT) and direct preference optimization (DPO).}
  \label{fig:training}
\end{figure}

The image encoder is a three-dimensional ViT-L\cite{vit} with 24 blocks, an embedding dimension of 1,024, four register tokens\cite{registers} and 302.9 million parameters (Fig.~\hyperref[fig:training]{\ref*{fig:training}a}). Each image token covers 2 $\times$ 16 $\times$ 16 voxels, corresponding to 3 $\times$ 8 $\times$ 8 mm. Positions are encoded in millimetres using three-dimensional rotary embeddings\cite{rope}. We trained the encoder from scratch without labels using self-distillation\cite{dinov2}, combining the DINO\cite{dino}, iBOT\cite{ibot} and KoLeo\cite{koleo} objectives with a cross-crop patch term. All stages respected the patient-level split.

Each sample consisted of one reconstruction and yielded two global crops of 64 $\times$ 192 $\times$ 192 voxels and eight local crops of 32 $\times$ 96 $\times$ 96 voxels. The teacher received one global crop in the full intensity range and another in an intensity window sampled from the full range and the series' native windows. When a lung window was present, the second view was replaced with a 10 mm lung maximum-intensity projection with probability 0.25. The student received additional intensity perturbations and iBOT masking.

Training ran for 20,000 steps with a global batch size of 1,024 reconstructions on 128 NVIDIA H200 GPUs, followed by 5,000 steps with an additional Gram-anchoring loss\cite{dinov3}. This loss kept pairwise similarities between the student's patch features close to those of a frozen copy of the teacher from step 20,000. Blocks 1 to 23 of the teacher were frozen for all subsequent stages.

\subsection{Alignment with report sentences}

Before training the classifier, we trained block 24 with its final normalization layer, a 512-dimensional projection and a view-type embedding contrastively against the statements of presence and absence derived from each training report (Fig.~\hyperref[fig:training]{\ref*{fig:training}b}). The statements were encoded by a fine-tuned CXR-BERT\cite{cxrbert}, and the loss treated all statements of a report as positives. Training ran for ten epochs on 304,921 training reports, three with a cosine schedule and seven of a continuation at half the learning rate. This block initialized the trainable block of the classifier.

\subsection{Organ-routed multiple-instance classifier}

The classifier treats each examination as a bag of image tokens from all its views (Fig.~\hyperref[fig:training]{\ref*{fig:training}c} and Supplementary Fig.~\ref{fig:supp-architecture}). The frozen block-23 tokens pass through the trainable block 24 and its final layer norm. Its outputs are standardized with fixed statistics and extended with 24 Fourier features\cite{fourier} of the token position in the scan grid. For each finding, a gated attention module\cite{ilse2018} with hidden size 512 scores every token, and a two-layer network with hidden size 384 and a GELU\cite{gelu} assigns every token a logit. The attention scores are normalized over all tokens of the bag, and the study logit is the attention-weighted mean of the token logits with a per-finding scale and bias.

Each finding was routed to its anatomical region. We derived 63 regions from CADS\cite{cads} organ segmentations, assigned a token to a region when the region covered at least 25\% of the token, and defined the route of a finding as the union of the regions of the finding and its descendants. Tokens outside the route receive a learned penalty, the softplus of a per-finding parameter initialized at 5, subtracted from their attention score. A finding therefore attends mainly to its organs while every token remains available, and a view with fewer than eight tokens in the route is treated as entirely inside it. Thirty findings were left unrouted, including those whose organ may be displaced or absent, such as ectopic kidney.

The classifier was trained with binary cross-entropy summed over the labelled findings and divided by the number of findings, without positive weighting, on one study bag per GPU and 128 bags per step, with token dropout of 0.2 and feature dropout of 0.1. Training ran in three stages of 38,740 steps in total as the routing table was refined, and at the start of the last stage the route penalty was reset to 1 for the 79 routed findings with fewer than 50 training positives. The final model is the checkpoint at the end of the last stage (epoch 4), with raw rather than exponential-moving-average weights. Per-finding Platt scaling\cite{platt1999}, fitted on training predictions and regularized towards the identity, converts logits into probabilities. The per-finding Youden\cite{youden1950} cut-off on training predictions is the decision threshold for classification metrics. Neither was refitted on validation or external data.

The signed evidence of a token for a finding is its attention-weighted contribution to the study logit. Evidence maps use its positive part, take the maximum over views in each cell of a world grid, are smoothed with a 4 mm Gaussian kernel and are restricted to the anatomical family of the finding, an organ, the lungs or lymph-node stations, when that family holds evidence. The smallest set of cells holding half of the smoothed evidence mass is split into connected components, and up to four of them, ranked by evidence mass, form the foci, each located at its mass-weighted centroid.

\subsection{Report generation}

For each examination, the report plan lists confident findings, with a calibrated probability of at least 0.5. It then includes possible findings at or above the training Youden cut-off, keeping the 20 with the highest calibrated probability. Each planned finding carries its calibrated probability relative to the training base rate, the distribution and sites of its evidence foci, located with the CADS segmentations, and an automated measurement where one exists. The writer is Gemma-4-31B-it\cite{gemma4} with LoRA\cite{lora} adapters of rank 16 and alpha 32 on the attention and feed-forward projections of its language model (Fig.~\hyperref[fig:training]{\ref*{fig:training}d}). It receives the instruction and the plan as one user turn without a system prompt, the study bag compressed to 512 latent tokens by a query resampler\cite{flamingo} together with study-type tokens, a pooled token and up to 16 focus tokens of evidence for each planned finding, and up to six zoomed evidence images per finding (Supplementary Figs.~\hyperref[fig:supp-reportgen-labels]{\ref*{fig:supp-reportgen-labels}a} and~\ref{fig:supp-prompts-writers}). These are axial, coronal and sagittal views of the main site and axial views of up to three further sites, each encoded by the Gemma-4 vision tower and pooled to one token.

The writer was trained to produce the Findings section of the reference report in two supervised stages, 4,666 steps over 295,541 training reports and a further 1,750 steps with the final plan tiers, followed by 300 steps of direct preference optimization\cite{dpo} (Supplementary Fig.~\hyperref[fig:supp-reportgen-labels]{\ref*{fig:supp-reportgen-labels}b}). The vision tower, connector and LoRA adapters were trained in every stage. Preference optimization used the sigmoid loss on summed sequence log-probabilities with $\beta$ = 0.1, two pairs per study and the frozen second-stage writer as reference, together with a cross-entropy term on the preferred report. Pairs were drawn from a pool of 1,636,197 pairs from 290,033 training reports, of which about 72,000 were used. Each pair contrasted the reference with a version that omitted one or several findings, added a finding from the plan or at random, replaced abnormal statements with normal ones, repeated statements, changed an attribute such as side, size or severity, or asserted a finding that the reference explicitly denied. Reports were decoded by nucleus sampling\cite{nucleus} with temperature 0.8, top-p 0.9 and a per-study seed.

\subsection{Chest radiograph model}

The chest radiograph model follows the same recipe. Its encoder was trained by the same self-distillation\cite{raydino}, without the Gram-anchoring phase, on 272,183 radiographs of the training split, each treated as a two-slice volume at native resolution, for 30,000 steps, and block 24 was aligned with report statements for ten epochs. Labels came from the same three-pass extraction with a vocabulary of 311 findings (Supplementary Fig.~\hyperref[fig:supp-prompts-labels]{\ref*{fig:supp-prompts-labels}c}), and the 129 findings with at least 50 training positives became classifier targets. The classifier adds a learned view embedding for frontal, lateral and other films to the tokens before block 24 and uses no routing. It was trained on 257,595 studies from 173,864 patients, and 28,965 studies from 19,802 patients were held out by patient. The signed evidence of each finding is measured relative to its mean token level in the studies negative for the finding among 2,000 sampled training studies. The writer was trained by supervised fine-tuning, preference optimization and a second round of preference optimization on edited versions of its own reports, and reports were decoded greedily.

\subsection{Comparison models}

We compared FORA with 18 released CT models, namely CoLiPri\cite{colipri}, CT-CLIP, CT-Vocab and CT-LiPro\cite{ctclip}, fVLM\cite{fvlm}, Jolia\cite{jolia}, Merlin\cite{merlin}, Pillar-0\cite{pillar0}, RADAR\cite{radar}, EXACT\cite{exact} in its pretrained and fine-tuned versions, ClinFusion-8B and ClinFusion-32B\cite{clinfusion}, NV-Reason-CT\cite{nvreasonct}, CT-CHAT\cite{ctclip}, Med3DVLM\cite{med3dvlm}, RadFM\cite{radfm} and MedGemma\cite{medgemma}, and additionally with M3D\cite{m3d} on the public benchmarks. All models were evaluated using their released weights, without retraining or probes. For contrastive models, we calculated the cosine margin between presence and absence prompts averaged over reconstructions, and used fixed prompt pairs for Merlin and Pillar-0. Generative models were given the longest series of each examination and were asked about each finding with their native prompt template (Supplementary Fig.~\hyperref[fig:supp-prompts-judge]{\ref*{fig:supp-prompts-judge}b}). Their decisions were the parsed yes or no answers, and ranking metrics used the next-token probability of yes or the margin between yes and no where the model exposed it. RADAR was scored with the released probability of each finding, averaged over reconstructions, and its native threshold of 0.5. We matched its 146 finding outputs to 202 of our 1,028 findings and to 9 of 18, 24 of 30 and 9 of 16 findings on CT-RATE, the Merlin test set and RAD-ChestCT, taking the highest probability when several outputs matched one finding and using its released protocol on the Merlin test set. Its released code does not process volumes with more than 1,000 voxels along an axis after resampling, which applied to 1,585 held-out examinations. Findings that a model cannot output and examinations that it cannot process were counted as negative decisions, so aggregate results reflect both coverage and performance.

For report generation we compared the seven CT report generators whose released inference ran on our data, namely NV-Reason-CT, CT-CHAT 70B and 8B, Merlin, BTB3D\cite{btb3d}, Med3DVLM and RadFM, each with its authors' prompt, decoding and token limit. We excluded MedGemma, which has no published CT report-generation prompt, ClinFusion, whose report protocol uses indications derived from the reference, CT-SSG, whose code is not released, EXACT-CHAT, whose checkpoint is incompatible with the released image export, and Astra, whose released inference failed on our data. For chest radiographs we compared seven classifiers (RadZero\cite{radzero}, CARZero\cite{carzero}, KAD\cite{kad}, MedKLIP\cite{medklip}, BioViL-T\cite{biovilt}, GLoRIA\cite{gloria} and MedGemma) and six report generators (LLaVA-Rad\cite{llavarad}, CheXagent\cite{chexagent}, Libra\cite{libra}, CXRMate\cite{cxrmate}, RaDialog\cite{radialog} and MedGemma).

\subsection{Evaluation of finding recognition}

We report AUROC, AUPRC, balanced accuracy and CRG\cite{crg}. Balanced accuracy and CRG use each model's own operating point, the training Youden cut-off for FORA and the native decision rule for the other models. On the held-out set, AUROC, AUPRC, and balanced accuracy were averaged over all pairs of finding and examination type with at least 10 positive and 50 negative examinations (2,132 pairs covering 854 findings), and CRG pooled all decisions. Confidence intervals came from a first-order linearization of each metric with 200,000 patient-level multiplier draws\cite{efron1993}, and FORA was compared with each model by two-sided paired tests with Holm correction\cite{holm1979} across the 72 internal comparisons.

On the public benchmarks, each model was scored on the full finding set of the benchmark, 18 findings on CT-RATE\cite{ctclip} (1,564 studies from 1,304 patients), 30 on the Merlin test set\cite{merlin} (all 5,125 test studies of its public release) and 16 on RAD-ChestCT\cite{radchestct} (3,630 scans). FORA used the same final checkpoint as on the held-out set without refitting, with its raw weights on the Merlin test set and RAD-ChestCT and its exponential-moving-average weights on CT-RATE (macro AUROC difference from the raw weights below 0.0001). For NV-Reason-CT on CT-RATE, the Results and Fig.~\hyperref[fig:classification]{\ref*{fig:classification}c} give the macro AUROC of 0.871, where each volume is a separate input. Intervals came from 200,000 paired bootstrap draws, resampling patients on CT-RATE and studies on the other sets, with Holm correction across 186 comparisons. For renal cysts in KiTS23\cite{kits21} (489 scans), rib fractures in RibFrac\cite{ribfrac} (80 scans) and the 21 Merlin findings covered by RADAR's released protocol, intervals came from 2,000 case-bootstrap resamples, and FORA was compared with the strongest alternative by a two-sided normal approximation with Holm correction across the three comparisons. Calibration of the Platt-scaled probabilities was assessed for findings with at least 10 validation positives, 900 for CT and 99 for chest radiographs, with reliability curves and the expected calibration error\cite{guo2017} over 15 equal-count bins (Supplementary Fig.~\ref{fig:supp-calibration}). Subgroup analyses stratified the held-out set by age, sex, scanner manufacturer and tube voltage (Supplementary Fig.~\ref{fig:supp-subgroups}).

\subsection{Evaluation of localization}

Localization used FORA's evidence maps without any fitting to the annotations. The top-focus hit rate is the proportion of annotated instances for which the centroid of the highest-mass evidence focus lies inside the annotation. Evidence overlap is the proportion of instances whose annotation intersects the smallest set of 3 $\times$ 8 $\times$ 8 mm cells holding 50\% or 80\% of the smoothed evidence mass. Chance for the hit rate is the probability that a uniformly sampled token centre within the route of the finding lies inside the annotation. We used FUMPE\cite{fumpe} (35 scans, 33 with emboli), KiTS23\cite{kits21} (489 scans with renal masses, 248 of them with cysts), RibFrac\cite{ribfrac} (80 scans, 60 with fractures) and ReXGroundingCT\cite{rexgrounding} (414 CT-RATE validation studies with 1,139 evaluable instances in 11 categories). Intervals for the hit rate are exact binomial intervals\cite{clopper1934} for FUMPE, KiTS23 and RibFrac. For ReXGroundingCT they come from 20,000 study-bootstrap resamples, and from 2,000 for the analysis by annotated volume.

We compared these maps with RADAR and GLINT-CT\cite{glint} on every annotated target of the five settings, namely pulmonary emboli in FUMPE, renal cysts and renal masses in KiTS23, rib fractures in RibFrac and all 1,139 evaluable ReXGroundingCT findings, with lung nodules (305 instances in 211 studies) and pleural effusions (28 in 22) also shown on their own. A model without output for a target scored 0 on every measure for that target. RADAR was run with its released checkpoint and inference code. Its evidence was the attention map that RADAR provides as visual evidence for a finding, the weights of its organ query over the image tokens at three scales, taken from the forward pass that produces its released probability. The map belongs to the organ query, so the kidney map served renal cysts and masses, the rib map rib fractures and the lung map the three ReXGroundingCT categories covered by RADAR's findings, lung nodule, lung opacity and pleural effusion, matched as in the classification comparison. RADAR has no prompt for pulmonary embolism or for the other eight ReXGroundingCT categories. The map was reduced to the 3 $\times$ 8 $\times$ 8 mm cells by the maximum per cell and scored with the code used for FORA. Because RADAR's 80\% evidence regions covered on average 36 to 212 times more cells than FORA's, we compared evidence overlap at equal size, keeping for each instance as many of RADAR's highest-valued cells as FORA's 80\% region contained and counting cells tied at the cut-off at their expected value. 

GLINT-CT was run with its released checkpoint. Its region was the part of its similarity map above the threshold that gave the best mean Dice on each set, chosen from 101 values between 0 and 1 as in its paper, and its first focus was the map-weighted centroid of the connected region with the largest summed map value. For overlap at equal size we used, for each instance, the highest threshold at which its region held at least as many cells as FORA's 80\% region. Paired differences used 1,000 bootstrap resamples of cases, or of studies for ReXGroundingCT, and the intervals of each model used 2,000.

For chest radiographs, the evidence of the classes mapped to each annotation was combined with weights given by their probabilities, and the hit rate uses the peak of the highest-mass focus, counted inside any box of the finding and averaged over findings. The variant of the map was chosen on separate box-annotated development sets, the training split of ChestX-Det10\cite{chestxdet10} and the training and validation splits of PadChest-GR\cite{padchestgr}. Test sets were the box-annotated subsets of ChestX-ray14\cite{chestxray14} (880 images), ChestX-Det10 (459), RSNA Pneumonia\cite{rsna} (1,218) and PadChest-GR (464).

\subsection{Evaluation of report generation}

Generated and reference Findings sections were labelled by the same three-pass Gemma-4-31B extractor with protocol-specific vocabularies (Supplementary Fig.~\hyperref[fig:supp-prompts-labels]{\ref*{fig:supp-prompts-labels}a,c}) and mapped to the 1,028 CT findings or the 129 chest radiograph findings. Asserted or uncertain findings counted as positive. Macro recall, precision and balanced accuracy were averaged over findings with at least 10 reference positives in each evaluated set, and CRG was pooled over the same findings. An LLM judge (Gemma-4-31B, Supplementary Fig.~\hyperref[fig:supp-prompts-judge]{\ref*{fig:supp-prompts-judge}a}) compared each generated report with the reference and counted matched findings and clinically significant errors of six kinds, namely missed findings, false findings, wrong location, wrong severity, false comparison and omitted comparison. The score of a report is the number of matched findings divided by the sum of matched findings and significant errors. A report with no matched finding scores 1 when it has no significant error and 0 otherwise. In the pairwise comparison, a separate prompt asked the judge to compare the reports of FORA and NV-Reason-CT with the reference in both presentation orders. Intervals came from 1,000 bootstrap resamples, clustered by patient on CT-RATE for the extractor metrics, and paired tests from 2,000 bootstrap resamples, clustered by patient on CT-RATE, with Holm correction within each figure panel.

The preference-optimization ablation compared, for CT, the first supervised writer with the same checkpoint after a separate DPO run of 375 steps that is not part of the released writer, on a fixed set of 500 validation studies. Both arms used greedy decoding, the first-stage plan and the training-label extractor, with micro-averaged metrics over findings with at least 10 reference positives. For chest radiographs it compared the supervised writer with the same writer after 200 DPO steps on all 28,965 validation studies. Within each comparison, plans and scoring were identical (Supplementary Fig.~\ref{fig:supp-dpo}).

\subsection{Cohort and representation analyses}

The longitudinal analyses of Extended Data Fig.~\ref{fig:extended-data-1} use report labels only. For each patient and finding, the anchor was the first date with an asserted or uncertain label in a single examination type, and report states at the next dates of that type were compared, with reports on the same day combined by the precedence asserted, uncertain, denied and unmentioned.

Profile groups were obtained from the 1,028 per-finding evidence values of FORA before its final affine layer. These were standardized and reduced to 50 principal components fitted on training studies, normalized to unit length and clustered by k-means\cite{lloyd1982} fitted on training studies (Extended Data Figs.~\ref{fig:fora-ed2-global-atlas} and~\ref{fig:fora-ed3-regional-atlas}). This gave 16 groups for the whole cohort, from 50,000 randomly sampled training studies, and six groups within each of six CT categories (head, neck, chest, cardiac, abdomen and spine). Enrichment of reported findings in each group was tested on the earliest examination of each patient, with two-sided Cochran-Mantel-Haenszel tests\cite{mantel1959} stratified by CT category for the whole cohort and one-sided Fisher exact tests within categories, with Benjamini-Hochberg correction\cite{benjamini1995}.

For the tissue readouts, we averaged the frozen block-23 patch tokens of the image encoder over the tokens that the CADS segmentation assigns to an organ, using the soft-tissue window of the series with the most body tokens (Supplementary Fig.~\ref{fig:fora-supp4-quantitative-information}). Liver attenuation was read from the liver embedding, L1 trabecular attenuation and L3 muscle attenuation from the bone embedding, and aortic calcification from the abdominal-vessel embedding. Embeddings were normalized to unit length, standardized within each training fold and mapped to the measurement by ridge regression\cite{hoerl1970} with the penalty chosen among 13 values from 0.01 to 10,000. Five-fold cross-validation grouped by patient gave out-of-fold readouts for about 12,700 examinations of about 10,800 patients from both the training and the held-out sets. These examinations were recorded as non-contrast, with a median aortic blood attenuation below 70 HU and a median subcutaneous fat attenuation between $-$140 and $-$50 HU. Control readouts used age, sex and tube voltage, or the embedding of another organ. Intervals came from patient-cluster bootstrap resamples, 300 for the correlations with the measurements and 1,000 for the within-patient changes.

Repeat consistency was measured on held-out patients only (Supplementary Fig.~\ref{fig:fora-supp5-repeat-consistency}). For each patient and CT category, the first two examinations on distinct days formed a pair, and each repeat was matched to 10 to 20 examinations of other held-out patients with the same examination type, site, contrast, scanner manufacturer and sex and an age within five years. Similarity was the cosine between unit-length finding profiles in the 50-component space, and intervals came from 1,000 patient bootstrap resamples. All tests were two-sided unless stated otherwise.

\section*{Data availability}

\ifnaturesubmission
\else
The FORA dataset is not yet fully available. Researchers can request access at \href{mailto:contact@forithmus.com}{contact@forithmus.com}. To support independent benchmarking, evaluation on the complete held-out FORA test set is available through \url{https://research.forithmus.com/}. Researchers can submit their models as Docker containers for evaluation without downloading the underlying data. The chest radiograph dataset used to develop and evaluate the initial architecture is publicly available at \url{https://huggingface.co/datasets/Forithmus/FORA}.
\fi

\section*{Code availability}
The trained FORA models and code are publicly available at \url{https://github.com/forithmus/FORA}. FORA can also be accessed programmatically through an API at \url{https://api.forithmus.com/} and explored interactively in Litera (\url{https://litera.forithmus.com/}), an imaging workspace for viewing examinations, reviewing AI-generated report drafts and inspecting the spatial evidence behind individual findings.

\section*{Author contributions}
S.E., I.E.H., B.M. and B.K. designed the study and developed the research methodology. S.E. and U.T. performed dataset retrieval, and S.E., U.T. and A.G. carried out dataset preparation and preprocessing. U.T. and O.F.D. prepared the CADS anatomical segmentation labels for the dataset. S.E. implemented the FORA architecture, developed the model training and evaluation pipelines, conducted the computational experiments, performed statistical analyses, interpreted the results and prepared the figures. S.E., I.E.H., B.M., and B.K. coordinated the study, interpreted the findings and drafted the manuscript. U.T., A.G., J.Wa., X.W., H.R., S.S., J.Wo., T.S., M.X., M.K.O., D.Y., A.My., P.G., Y.H., C.Z., M.E., A.Ma., M.M. and D.X. provided domain-specific expertise, contributed to scientific discussions and provided feedback on the methodology and interpretation of the results. B.K., B.M. and I.E.H. jointly supervised the study. All authors reviewed and approved the final manuscript.

\section*{Acknowledgments}
High-performance computing resources were provided by the Erlangen National High Performance Computing Center (NHR@FAU) at Friedrich-Alexander-Universit\"at Erlangen-N\"urnberg under NHR projects b180dc, y100dc. NHR is funded by federal and Bavarian state authorities. The NHR@FAU hardware is partly funded by the German Research Foundation (DFG; grant 440719683). O.F.D. and B.K. are supported through the Bavarian AI Foundation Model Initiative, funded by the Bavarian State Ministry of Science and the Arts (StMWK). J.Wa. and B.K. are supported by ERC MIA-NORMAL 101083647, 101326871 and DFG 513220538, 512819079, 440719683, CL 552/1-1. 

\section*{Competing interests}
S.E., U.T., A.G. and I.E.H. are employees of Forithmus. D.Y., A.My., P.G., Y.H., C.Z. and D.X. are employees of NVIDIA, which developed NV-Reason-CT, one of the comparison models evaluated in this study. The remaining authors declare no competing interests.

\bibliographystyle{naturemag}
\bibliography{references}

\input{ed1-chest-xray-float}
\input{ed2-ct-classification-by-type-float}
\input{ed3-ct-localization-float}
\input{ed4-ct-report-generation-float}
\input{ed5-longitudinal-reporting-float}
\input{ed6-whole-cohort-atlas-float}
\input{ed7-regional-atlas-float}

\input{s1-data-preparation-ontology-float}
\input{s2-report-generation-labels-float}
\input{s3-prompts-ct-training-labels-float}
\input{s4-architecture-float}
\input{s5-evaluation-protocol-float}
\input{s6-classifier-breadth-training-data-float}
\input{s7-calibration-float}
\input{s8-external-transfer-float}
\input{s9-subgroup-robustness-float}
\input{s10-grounded-findings-gallery-float}
\input{s11-prompts-report-writers-float}
\input{s12-dpo-ablation-float}
\input{s14-prompts-judge-baselines-float}
\input{s15-report-examples-float}
\input{s16-quantitative-tissue-float}
\input{s17-repeat-consistency-float}

\end{document}

%% file: fig2-classification-localization-float.tex
\begin{figure}[p]
  \parbox[t][\textheight][t]{\textwidth}{%
    \vspace*{-5.2mm}%
    {\centering\includegraphics[width=\textwidth,height=\dimexpr\textheight+11.2mm-33.3mm\relax,keepaspectratio]{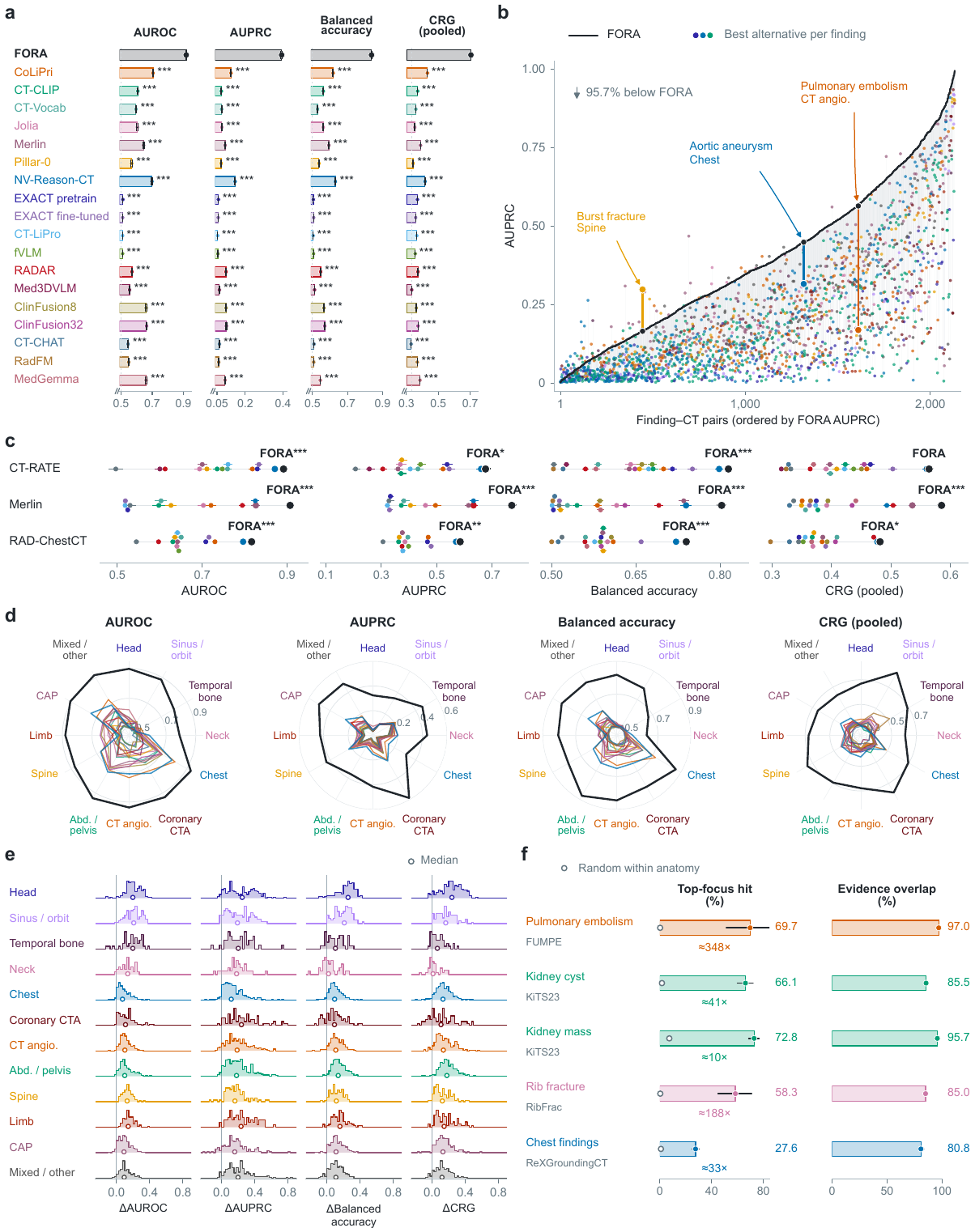}\par}
    \caption[CT finding recognition and localization.]{CT finding recognition and localization.
    \textbf{a}, FORA and 18 comparison models on held-out CT examinations.
    \textbf{b}, AUPRC versus the strongest alternative across 2,132 finding-CT-type pairs.
    \textbf{c}, Public benchmarks; NV-Reason-CT's CT-RATE AUROC is from its preprint.
    \textbf{d}, Performance across 12 CT types.
    \textbf{e}, AUROC advantage over the strongest alternative by CT type.
    \textbf{f}, Localization on four datasets; hollow points indicate chance.
    Error bars, 95\% confidence intervals. Significance from two-sided paired tests with Holm correction is marked $^{*}P<0.05$, $^{**}P<0.01$ and $^{***}P<0.001$.}
    \label{fig:classification}%
    \vspace*{-6.0mm}}
\end{figure}

%% file: fig3-report-generation-float.tex
\begin{figure}[p]
  \parbox[t][\textheight][t]{\textwidth}{%
    \vspace*{-5.2mm}%
    {\centering\includegraphics[width=\textwidth,height=\dimexpr\textheight+11.2mm-33.3mm\relax,keepaspectratio]{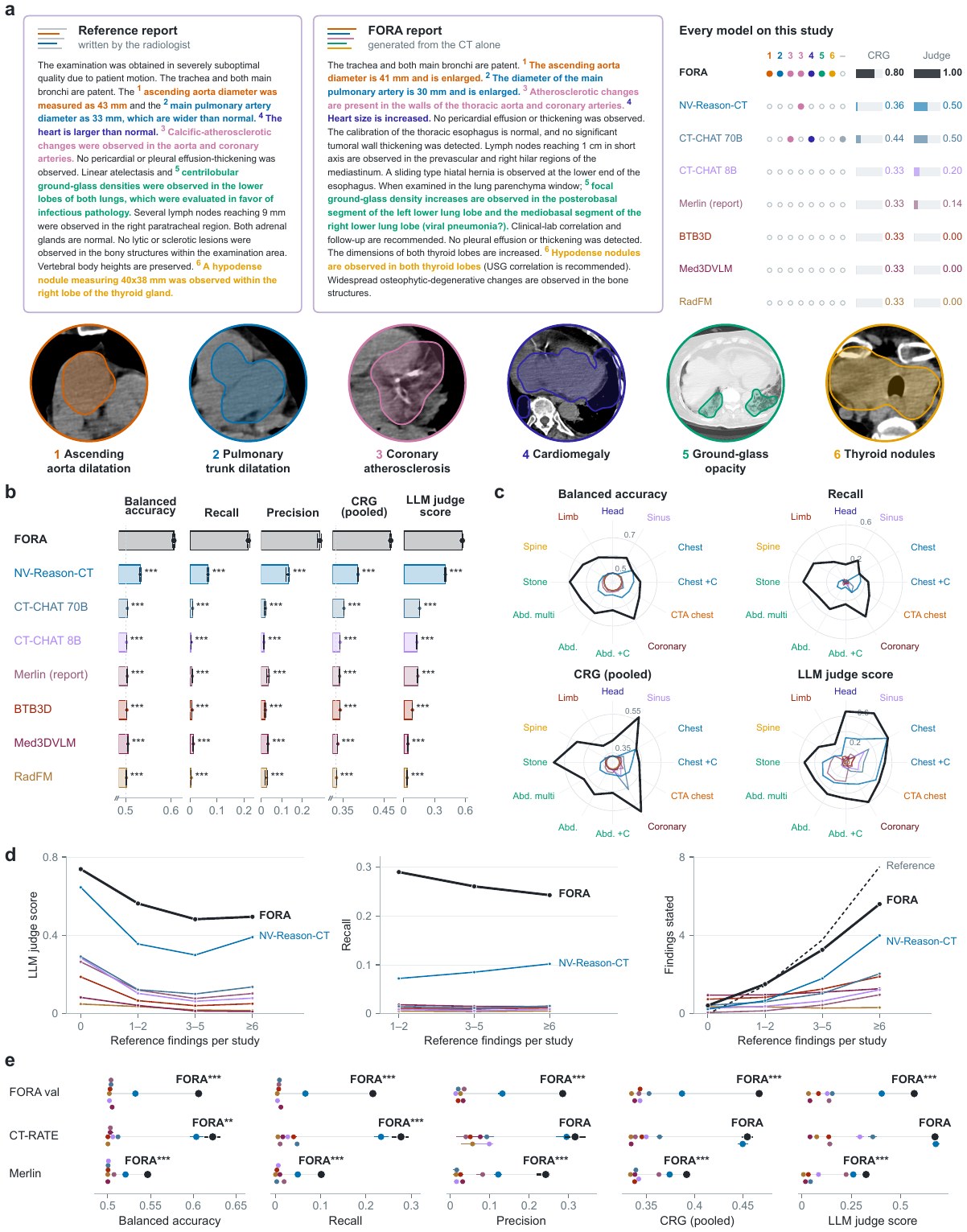}\par}
    \caption[Grounded CT report generation.]{Grounded CT report generation. \textbf{a}, Reference and FORA reports for a validation chest CT with matched statements numbered, and FORA's evidence. Dots mark the reference findings that each model's report recovered, with column numbers matching the statements. \textbf{b}, Finding-level metrics and LLM-judge score of FORA and seven other report generators on the FORA validation set. \textbf{c}, The 12 most frequent report protocols. \textbf{d}, Performance by number of reference findings per report. \textbf{e}, FORA validation set, CT-RATE and Merlin. Error bars, 95\% bootstrap confidence intervals. Stars, two-sided paired tests with Holm correction ($^{*}P<0.05$, $^{**}P<0.01$, $^{***}P<0.001$).}
    \label{fig:reportgen}%
    \vspace*{-6.0mm}}
\end{figure}

%% file: ed1-chest-xray-float.tex
\clearpage
\newgeometry{left=25mm,right=25mm,top=17.3mm,bottom=19.0mm,headsep=3.5mm,footskip=4.5mm}
\begingroup
\pagestyle{forithmus}
\setlength{\parskip}{0pt}
\makeatletter
\setlength{\@fptop}{0pt}
\setlength{\@fpbot}{0pt plus 1fil}
\makeatother
\begin{extendeddatafigure}[p]
  {\centering\includegraphics[width=\textwidth,height=\dimexpr\textheight-33.3mm\relax,keepaspectratio]{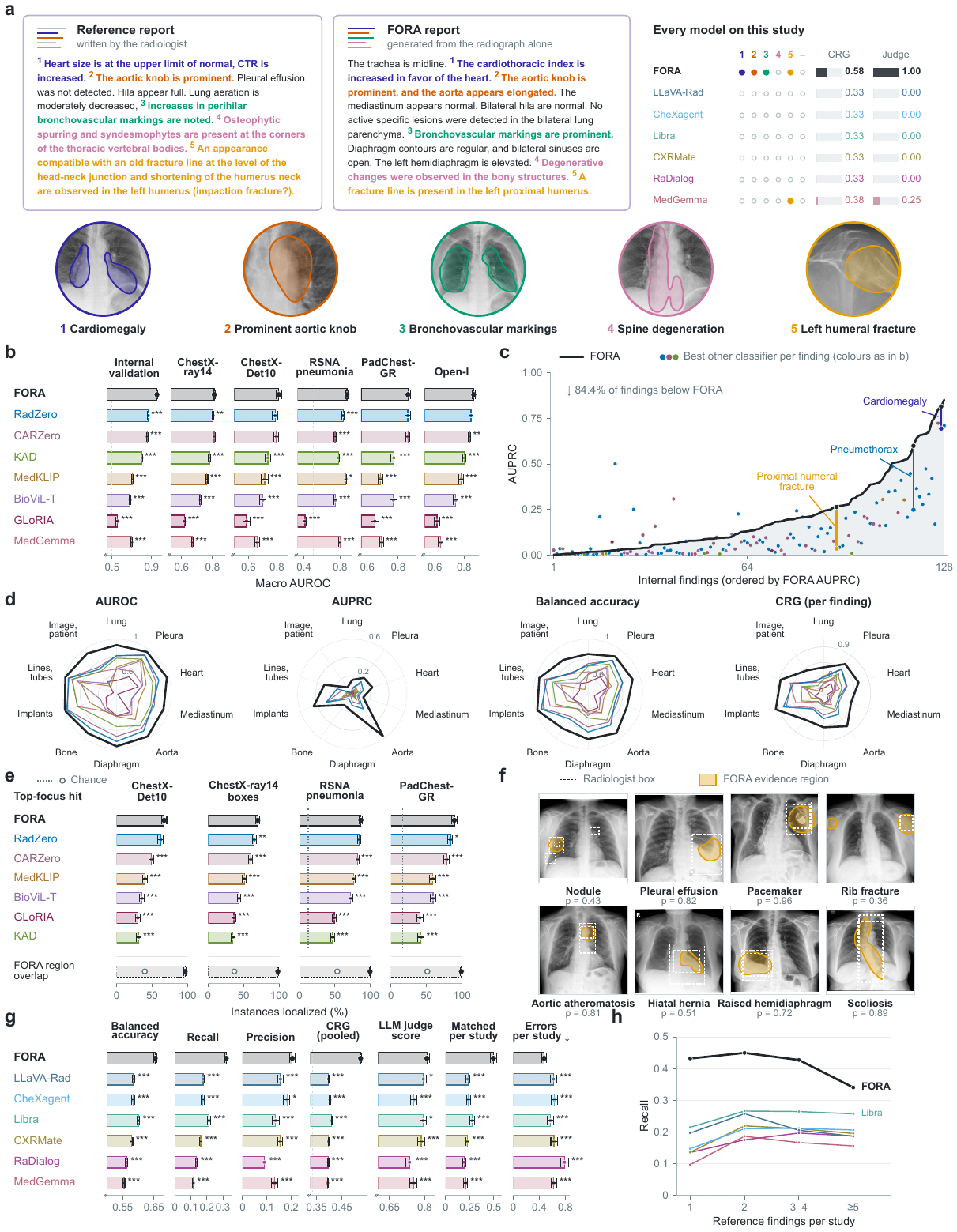}\par}
  \caption[FORA on chest radiographs.]{FORA on chest radiographs. \textbf{a}, Reference and FORA reports for an internal validation radiograph with matched statements numbered, and FORA's evidence. Dots mark the reference findings that each model's report recovered, with column numbers matching the statements. \textbf{b}, Macro AUROC on six test sets. \textbf{c}, AUPRC for 128 findings. \textbf{d}, Performance by finding category. \textbf{e}, Localization on four box-annotated test sets. \textbf{f}, Radiologist boxes and FORA's evidence on PadChest-GR. \textbf{g}, Report generation. \textbf{h}, Recall by number of reference findings. Error bars, 95\% bootstrap confidence intervals. Stars, two-sided paired tests with Holm correction ($^{*}P<0.05$, $^{**}P<0.01$, $^{***}P<0.001$).}
  \label{fig:ed-cxr}
\end{extendeddatafigure}
\clearpage
\endgroup
\restoregeometry
\pagestyle{forithmus}

%% file: ed2-ct-classification-by-type-float.tex
\clearpage
\newgeometry{left=25mm,right=25mm,top=17.3mm,bottom=19.0mm,headsep=3.5mm,footskip=4.5mm}
\begingroup
\pagestyle{forithmus}
\setlength{\parskip}{0pt}
\makeatletter
\setlength{\@fptop}{0pt}
\setlength{\@fpbot}{0pt plus 1fil}
\makeatother
\begin{extendeddatafigure}[p]
  {\centering\includegraphics[width=\textwidth,height=\dimexpr\textheight-25.5mm\relax,keepaspectratio]{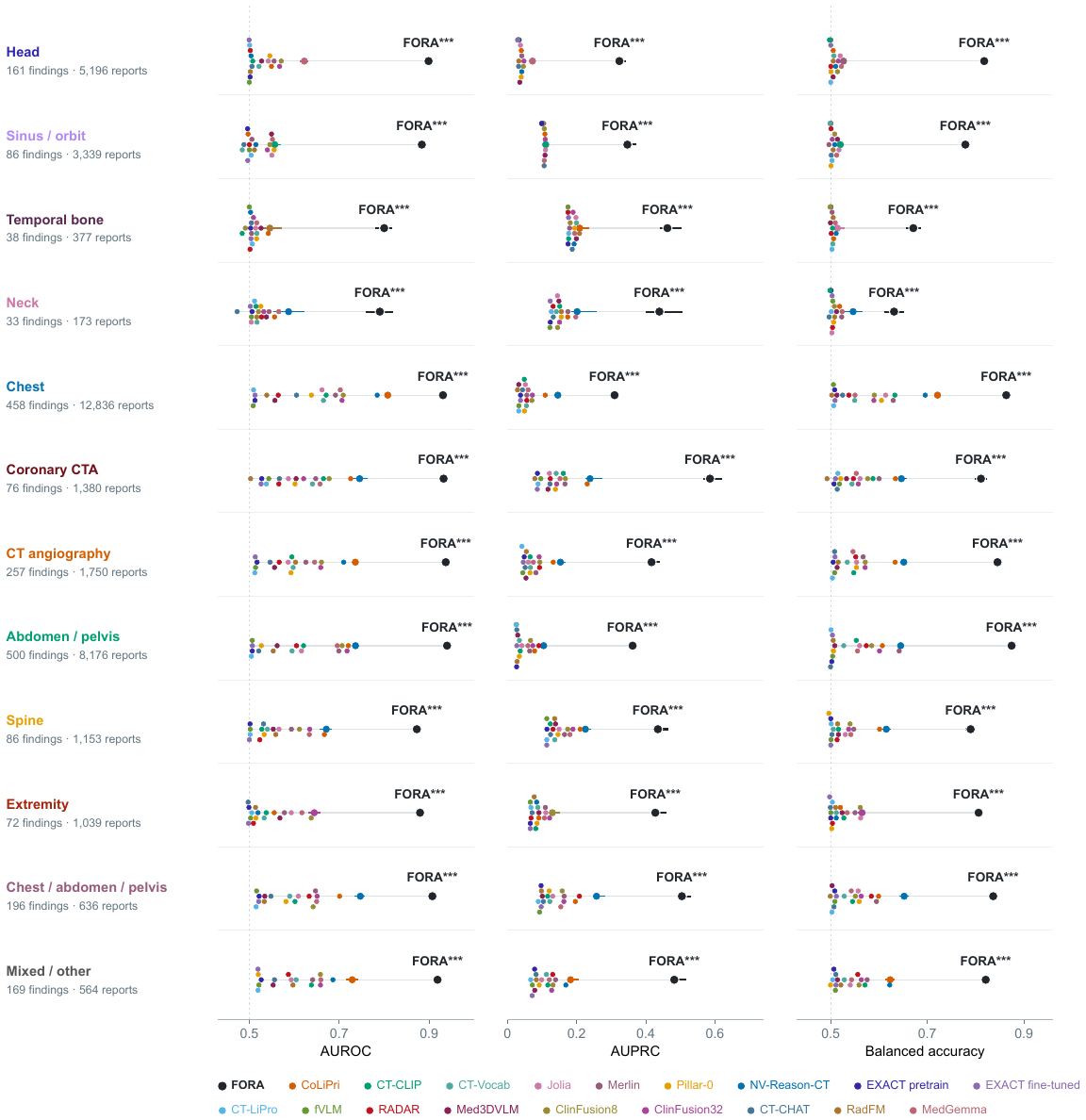}\par}
  \caption[CT finding classification in every CT type.]{CT finding classification in every CT type. Macro AUROC, AUPRC and balanced accuracy of FORA (black) and the 18 other models of Fig.~\ref{fig:classification} (coloured) in each CT type of the FORA validation set. The number of findings is given beside each type. Error bars, 95\% bootstrap confidence intervals. Stars, FORA against the best other model, two-sided paired tests with Holm correction ($^{***}P<0.001$).}
  \label{fig:ed-ct-types}
\end{extendeddatafigure}
\clearpage
\endgroup
\restoregeometry
\pagestyle{forithmus}

%% file: ed3-ct-localization-float.tex
\clearpage
\newgeometry{left=25mm,right=25mm,top=17.3mm,bottom=19.0mm,headsep=3.5mm,footskip=4.5mm}
\begingroup
\pagestyle{forithmus}
\setlength{\parskip}{0pt}
\makeatletter
\setlength{\@fptop}{0pt}
\setlength{\@fpbot}{0pt plus 1fil}
\makeatother
\begin{extendeddatafigure}[p]
  {\centering\includegraphics[width=\textwidth,height=\dimexpr\textheight-29.4mm\relax,keepaspectratio]{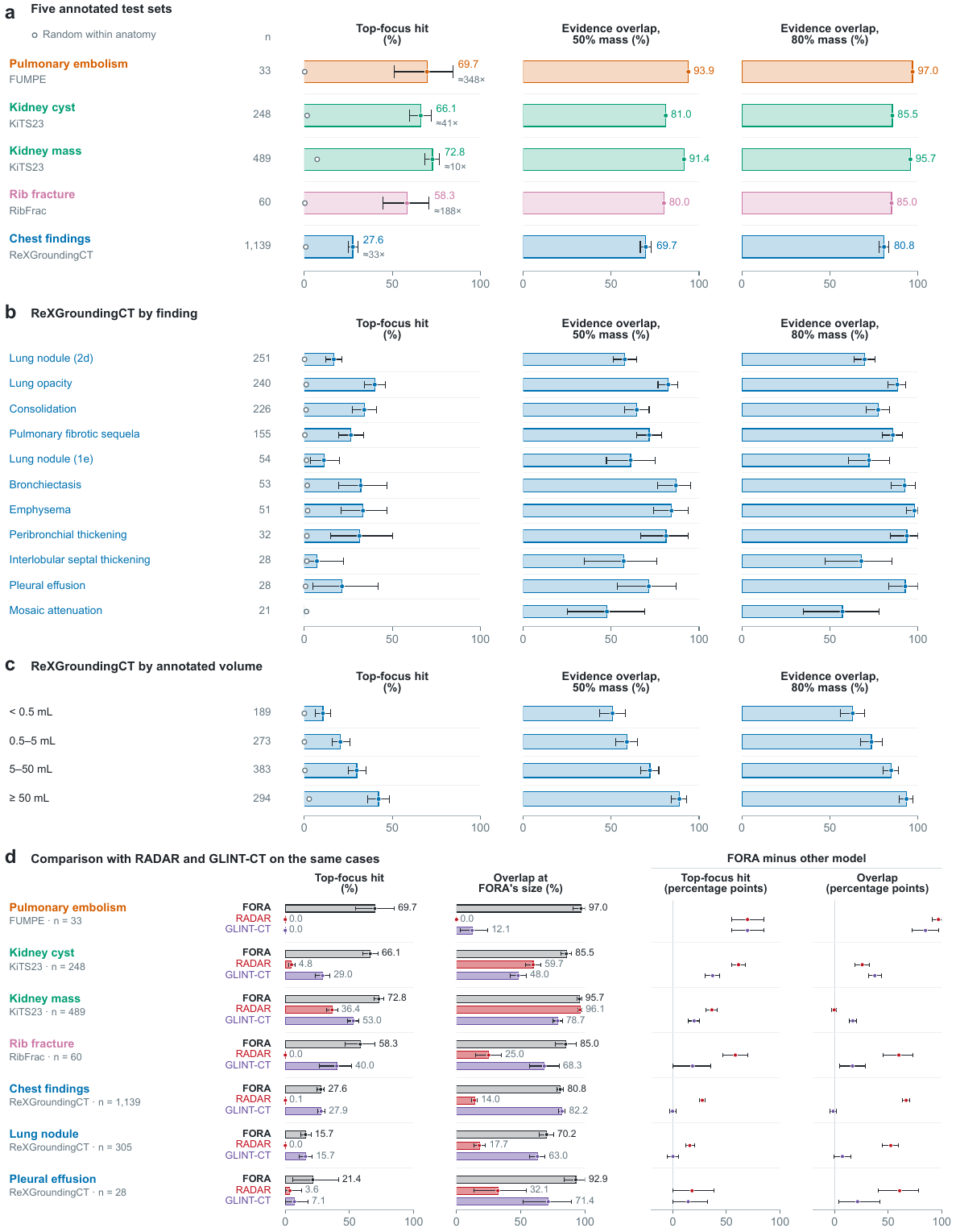}\par}
  \caption[CT localization from FORA's evidence.]{CT localization from FORA's evidence. \textbf{a}, Top-focus hit and evidence overlap in FUMPE, KiTS23, RibFrac and ReXGroundingCT. Hollow points, chance. \textbf{b}, Every ReXGroundingCT finding category. \textbf{c}, ReXGroundingCT findings by annotated volume. \textbf{d}, FORA, RADAR and GLINT-CT on the same annotated targets, with paired differences. A model with no output for a target scores 0 there, as RADAR does for pulmonary embolism and for 8 of the 11 ReXGroundingCT categories. For overlap, each other model's region is cut to the size of FORA's 80\% region. Error bars, 95\% confidence intervals.}
  \label{fig:ed-localization}
\end{extendeddatafigure}
\clearpage
\endgroup
\restoregeometry
\pagestyle{forithmus}

%% file: ed4-ct-report-generation-float.tex
\clearpage
\newgeometry{left=25mm,right=25mm,top=17.3mm,bottom=19.0mm,headsep=3.5mm,footskip=4.5mm}
\begingroup
\pagestyle{forithmus}
\setlength{\parskip}{0pt}
\makeatletter
\setlength{\@fptop}{0pt}
\setlength{\@fpbot}{0pt plus 1fil}
\makeatother
\begin{extendeddatafigure}[p]
  {\centering\includegraphics[width=\textwidth,height=\dimexpr\textheight-29.4mm\relax,keepaspectratio]{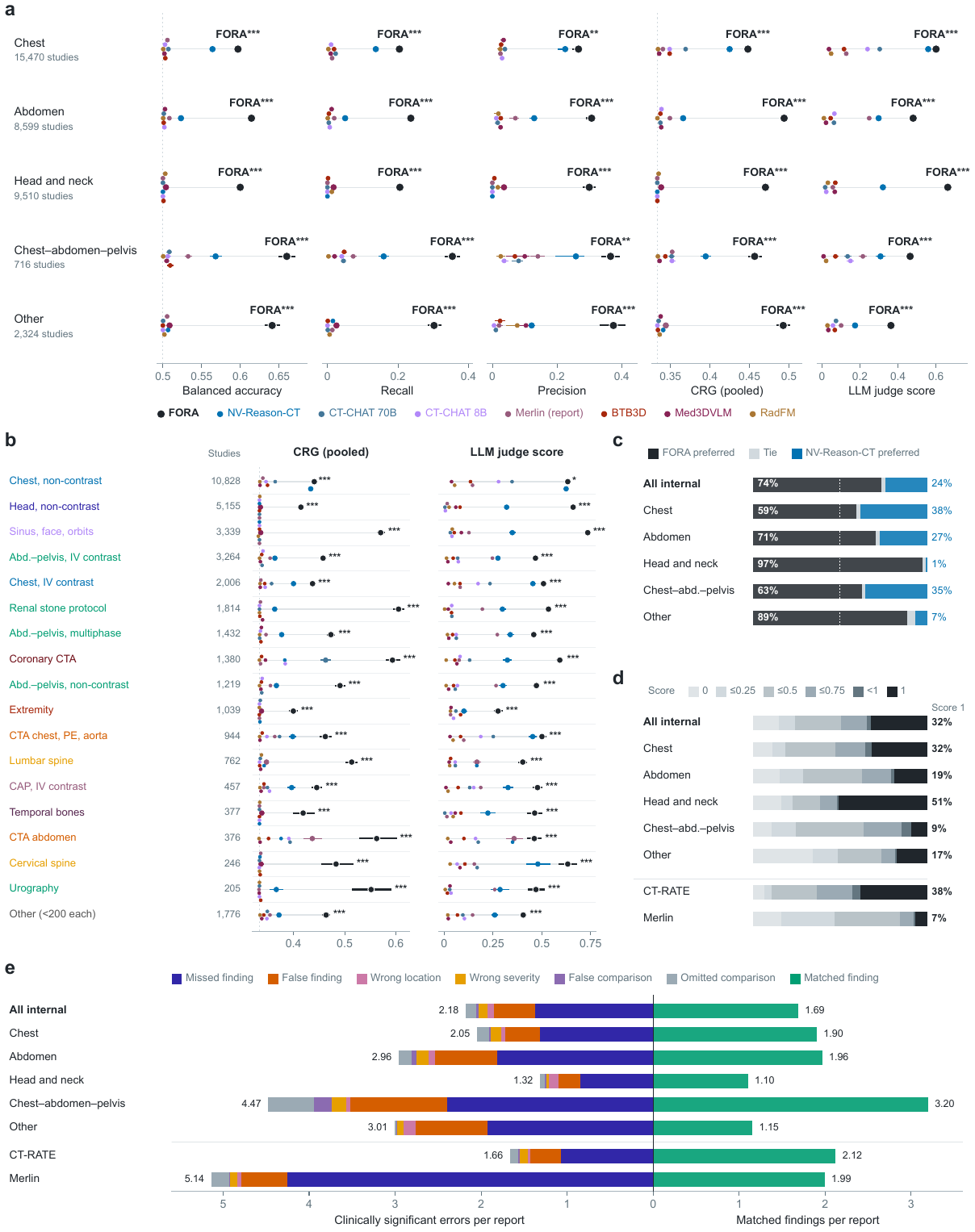}\par}
  \caption[CT report generation by body region, report protocol and error type.]{CT report generation by body region, report protocol and error type. \textbf{a}, Finding-level metrics and LLM-judge score of the eight report generators in each of five body regions of the FORA validation set. \textbf{b}, CRG score and judge score in 17 report protocols. \textbf{c}, Pairwise judge preference between FORA and NV-Reason-CT. \textbf{d}, Distribution of FORA's judge scores. \textbf{e}, Clinically significant errors and matched findings per FORA report. Error bars, 95\% bootstrap confidence intervals. Stars, two-sided paired tests with Holm correction ($^{*}P<0.05$, $^{**}P<0.01$, $^{***}P<0.001$).}
  \label{fig:ed-reportgen}
\end{extendeddatafigure}
\clearpage
\endgroup
\restoregeometry
\pagestyle{forithmus}

%% file: ed5-longitudinal-reporting-float.tex
\clearpage
\newgeometry{left=25mm,right=25mm,top=17.3mm,bottom=19.0mm,headsep=3.5mm,footskip=4.5mm}
\begingroup
\pagestyle{forithmus}
\setlength{\parskip}{0pt}
\makeatletter
\setlength{\@fptop}{0pt}
\setlength{\@fpbot}{0pt plus 1fil}
\makeatother
\begin{extendeddatafigure}[p]
  {\centering\includegraphics[width=\textwidth,height=\dimexpr\textheight-37.2mm\relax,keepaspectratio]{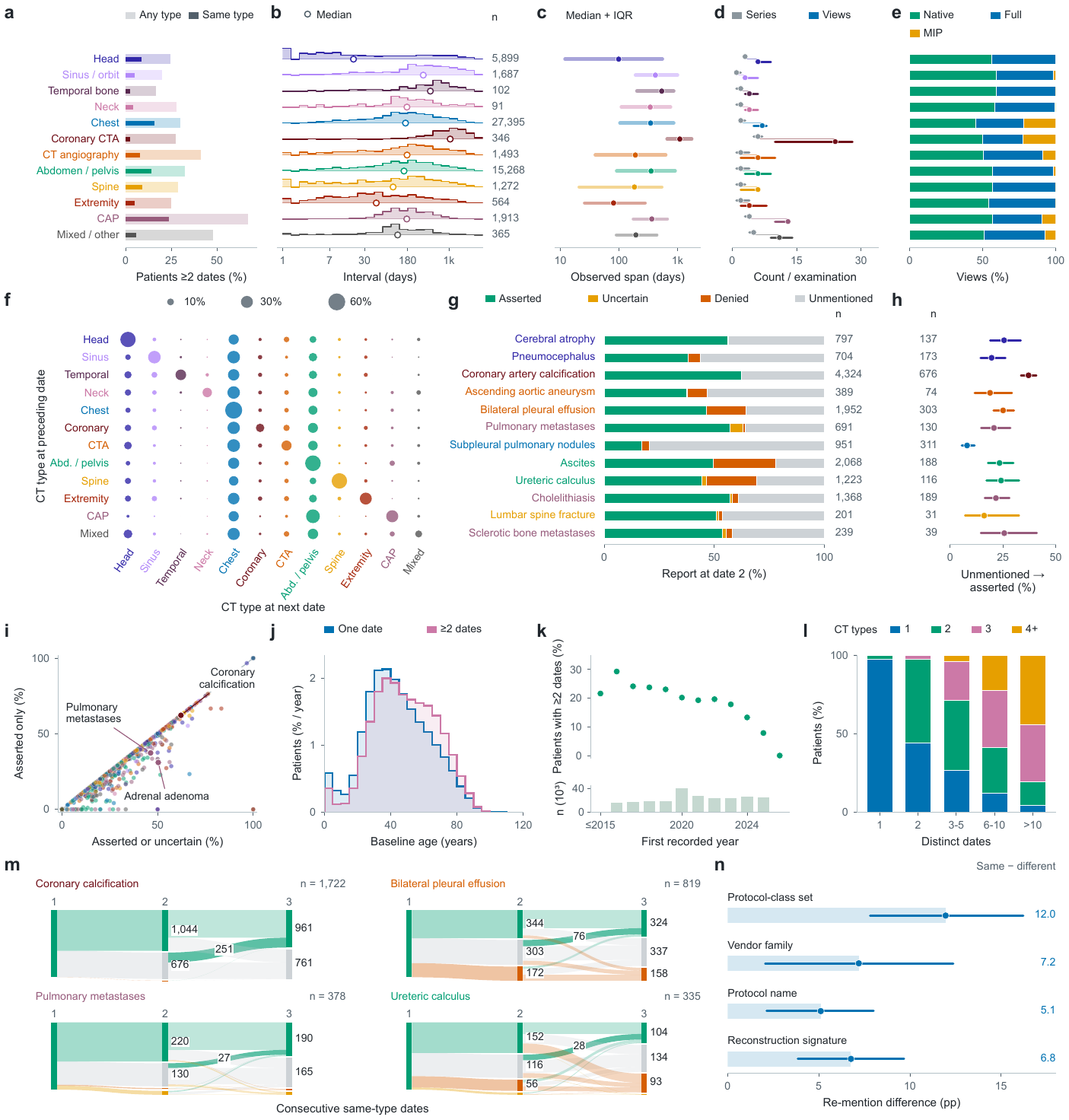}\par}
  \caption[Longitudinal observation and reporting in the FORA cohort.]{Longitudinal observation and reporting in the FORA cohort. \textbf{a}, Patients scanned on two or more dates. \textbf{b}, Intervals between consecutive dates. \textbf{c}, Follow-up span per patient. \textbf{d}, Series and views per examination. \textbf{e}, Native, full-range and lung MIP views. \textbf{f}, CT type at consecutive dates. \textbf{g}, Report state at the next date for 12 findings. \textbf{h}, Findings first asserted at the third date. \textbf{i}, Positive rates at the second date with and without uncertain labels. \textbf{j}, Age at the first date. \textbf{k}, Patients with several dates by first year. \textbf{l}, CT types per patient by number of dates. \textbf{m}, Three-date report trajectories for four findings. \textbf{n}, Repeated coronary calcification labels for identical and different acquisition settings. Error bars, 95\% confidence intervals. CT colours follow Fig.~\ref{fig:dataset}.}
  \label{fig:extended-data-1}
\end{extendeddatafigure}
\clearpage
\endgroup
\restoregeometry
\pagestyle{forithmus}

%% file: ed6-whole-cohort-atlas-float.tex
\clearpage
\newgeometry{left=25mm,right=25mm,top=17.3mm,bottom=19.0mm,headsep=3.5mm,footskip=4.5mm}
\begingroup
\pagestyle{forithmus}
\setlength{\parskip}{0pt}
\makeatletter
\setlength{\@fptop}{0pt}
\setlength{\@fpbot}{0pt plus 1fil}
\makeatother
\begin{extendeddatafigure}[p]
  {\centering\includegraphics[width=\textwidth,height=\dimexpr\textheight-21.6mm\relax,keepaspectratio]{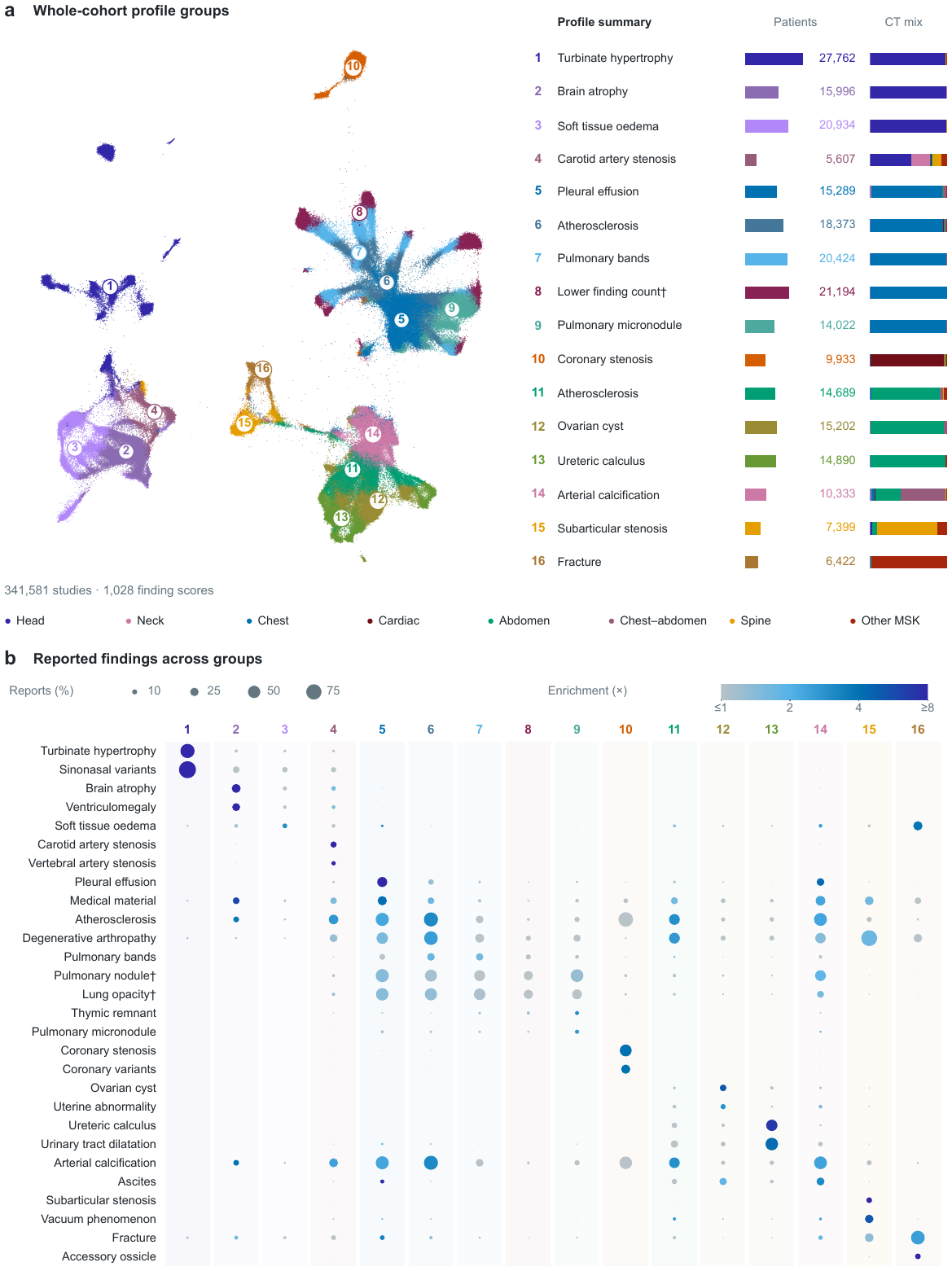}\par}
  \caption[Whole-cohort CT prediction-profile groups.]{Whole-cohort CT prediction-profile groups. \textbf{a}, UMAP of all 341,581 examinations coloured by 16 groups of FORA prediction profiles, with patients and CT types per group. \textbf{b}, Prevalence (dot area) and prevalence ratio (colour) of 28 reported findings in each group. Enrichment, CT-stratified Cochran--Mantel--Haenszel tests with Benjamini--Hochberg correction.}
  \label{fig:fora-ed2-global-atlas}
\end{extendeddatafigure}
\clearpage
\endgroup
\restoregeometry
\pagestyle{forithmus}

%% file: ed7-regional-atlas-float.tex
\clearpage
\newgeometry{left=25mm,right=25mm,top=17.3mm,bottom=19.0mm,headsep=3.5mm,footskip=4.5mm}
\begingroup
\pagestyle{forithmus}
\setlength{\parskip}{0pt}
\makeatletter
\setlength{\@fptop}{0pt}
\setlength{\@fpbot}{0pt plus 1fil}
\makeatother
\begin{extendeddatafigure}[p]
  {\centering\includegraphics[width=\textwidth,height=\dimexpr\textheight-21.6mm\relax,keepaspectratio]{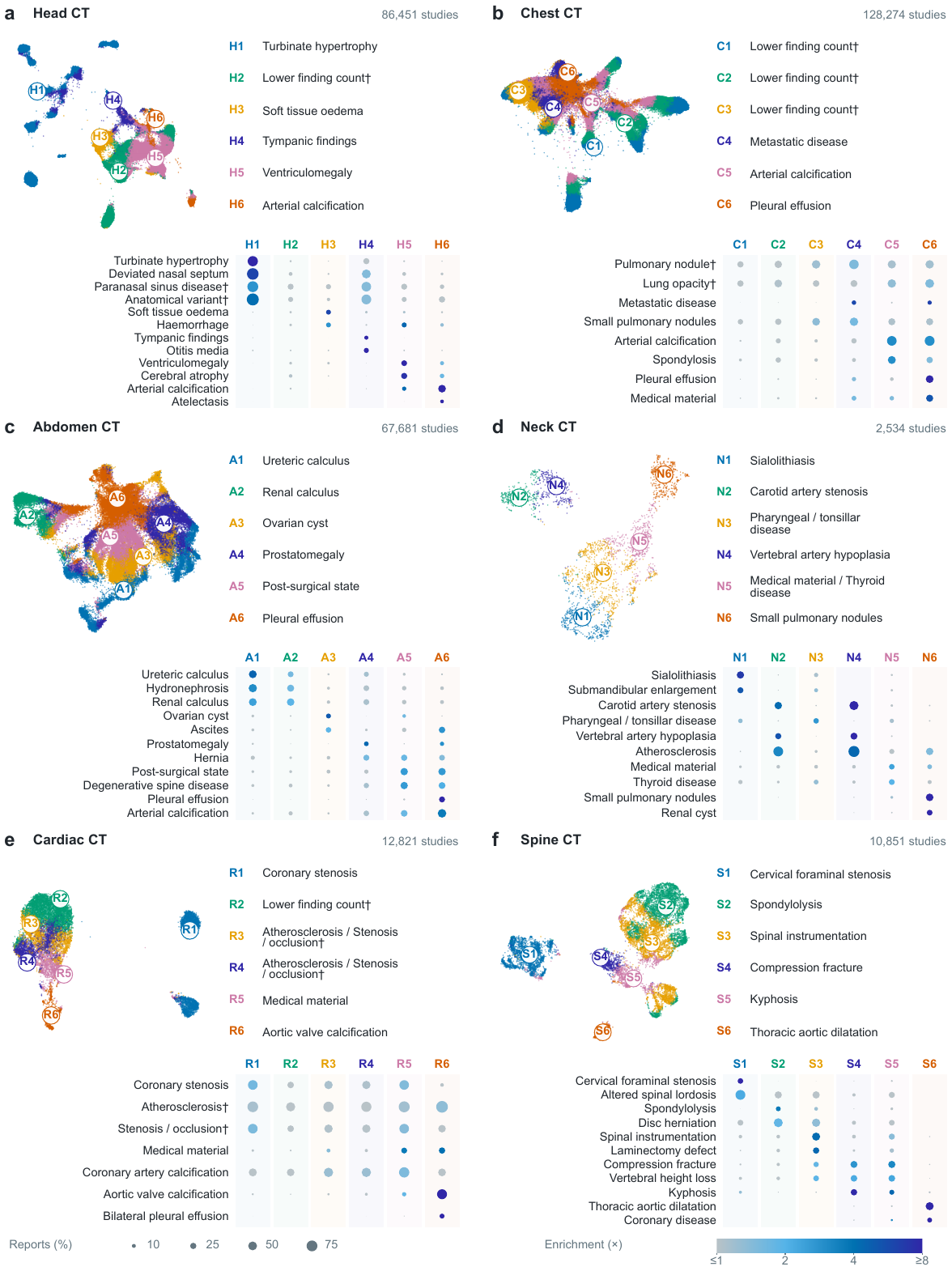}\par}
  \caption[Regional CT prediction-profile groups.]{Regional CT prediction-profile groups. \textbf{a--f}, UMAP of head, chest, abdominal, neck, cardiac and spine CT coloured by six prediction-profile groups, and prevalence (dot area) and prevalence ratio (colour) of reported findings in each group. Enrichment, one-sided tests with Benjamini--Hochberg correction.}
  \label{fig:fora-ed3-regional-atlas}
\end{extendeddatafigure}
\clearpage
\endgroup
\restoregeometry
\pagestyle{forithmus}

%% file: s1-data-preparation-ontology-float.tex
\clearpage
\newgeometry{left=25mm,right=25mm,top=17.3mm,bottom=19.0mm,headsep=3.5mm,footskip=4.5mm}
\begingroup
\pagestyle{forithmus}
\setlength{\parskip}{0pt}
\makeatletter
\setlength{\@fptop}{0pt}
\setlength{\@fpbot}{0pt plus 1fil}
\makeatother
\begin{supplementaryfigure}[p]
  {\centering\includegraphics[width=\textwidth,height=\dimexpr\textheight-21.6mm\relax,keepaspectratio,page=1]{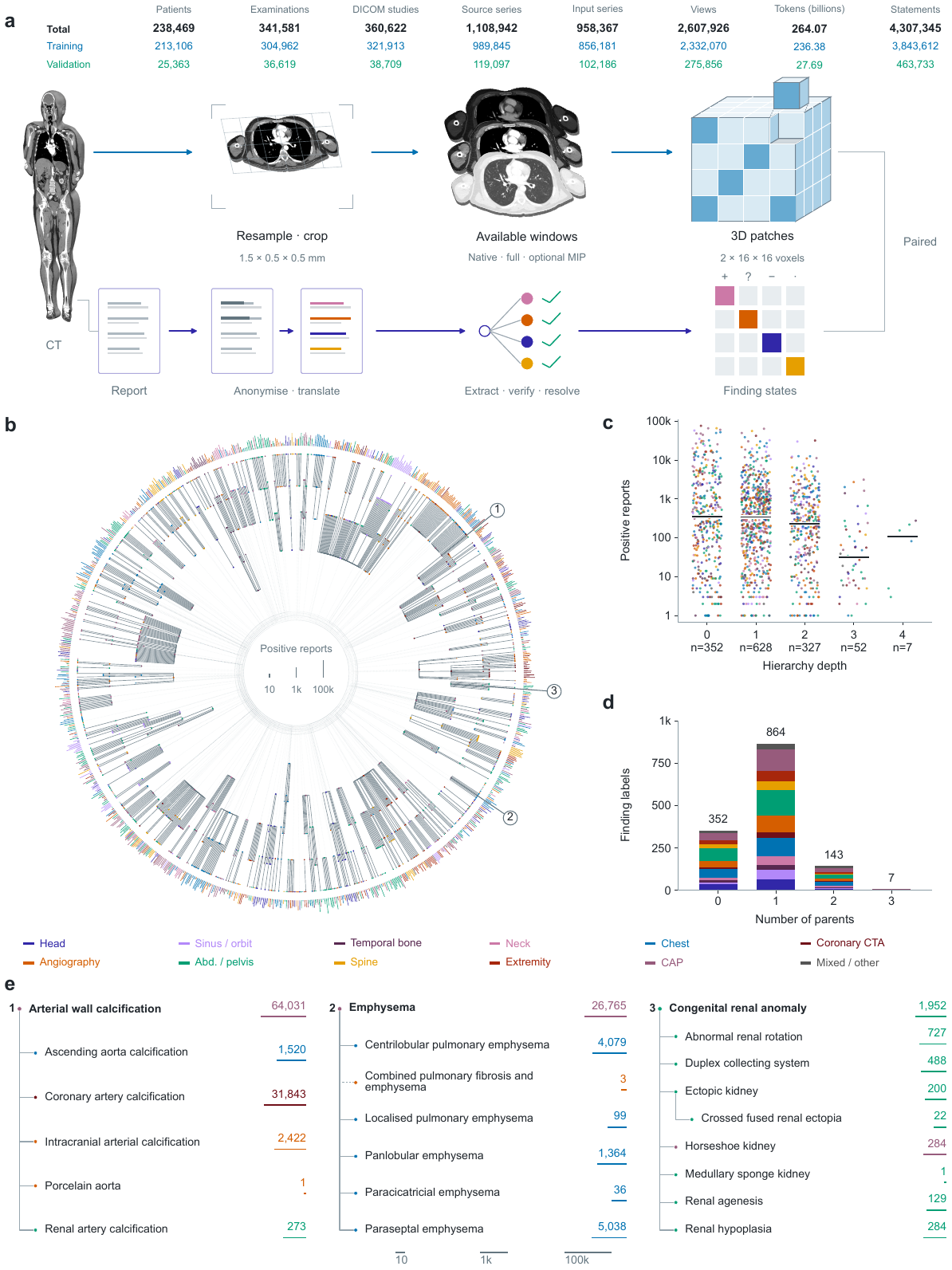}\par}
  \caption[Data preparation and finding ontology.]{Data preparation and finding ontology. \textbf{a}, CT and report preparation with patient-disjoint splits. \textbf{b}, Finding ontology of 1,366 labels. \textbf{c}, Positive reports by depth in the hierarchy. \textbf{d}, Labels by number of parents. \textbf{e}, Descendants of three example findings. Pages 2--5 list every label with its count and parents. Colours follow Fig.~\ref{fig:dataset}.}
  \label{fig:supplementary-1}
\end{supplementaryfigure}
\clearpage
\begin{supplementaryfigure}[p]
  \centering
  \includegraphics[width=\textwidth,height=\dimexpr\textheight-8mm\relax,keepaspectratio,page=2]{s1-data-preparation-ontology.pdf}\par\vspace{1mm}
  {\small\textsf{Supplementary Figure~\ref{fig:supplementary-1}, page 2 of 5}}
\end{supplementaryfigure}
\clearpage
\begin{supplementaryfigure}[p]
  \centering
  \includegraphics[width=\textwidth,height=\dimexpr\textheight-8mm\relax,keepaspectratio,page=3]{s1-data-preparation-ontology.pdf}\par\vspace{1mm}
  {\small\textsf{Supplementary Figure~\ref{fig:supplementary-1}, page 3 of 5}}
\end{supplementaryfigure}
\clearpage
\begin{supplementaryfigure}[p]
  \centering
  \includegraphics[width=\textwidth,height=\dimexpr\textheight-8mm\relax,keepaspectratio,page=4]{s1-data-preparation-ontology.pdf}\par\vspace{1mm}
  {\small\textsf{Supplementary Figure~\ref{fig:supplementary-1}, page 4 of 5}}
\end{supplementaryfigure}
\clearpage
\begin{supplementaryfigure}[p]
  \centering
  \includegraphics[width=\textwidth,height=\dimexpr\textheight-8mm\relax,keepaspectratio,page=5]{s1-data-preparation-ontology.pdf}\par\vspace{1mm}
  {\small\textsf{Supplementary Figure~\ref{fig:supplementary-1}, page 5 of 5}}
\end{supplementaryfigure}
\clearpage
\endgroup
\restoregeometry
\pagestyle{forithmus}

%% file: s2-report-generation-labels-float.tex
\clearpage
\newgeometry{left=25mm,right=25mm,top=17.3mm,bottom=19.0mm,headsep=3.5mm,footskip=4.5mm}
\begingroup
\pagestyle{forithmus}
\setlength{\parskip}{0pt}
\makeatletter
\setlength{\@fptop}{0pt}
\setlength{\@fpbot}{0pt plus 1fil}
\makeatother
\begin{supplementaryfigure}[p]
  {\centering\includegraphics[width=\textwidth,height=\dimexpr\textheight-13.8mm\relax,keepaspectratio]{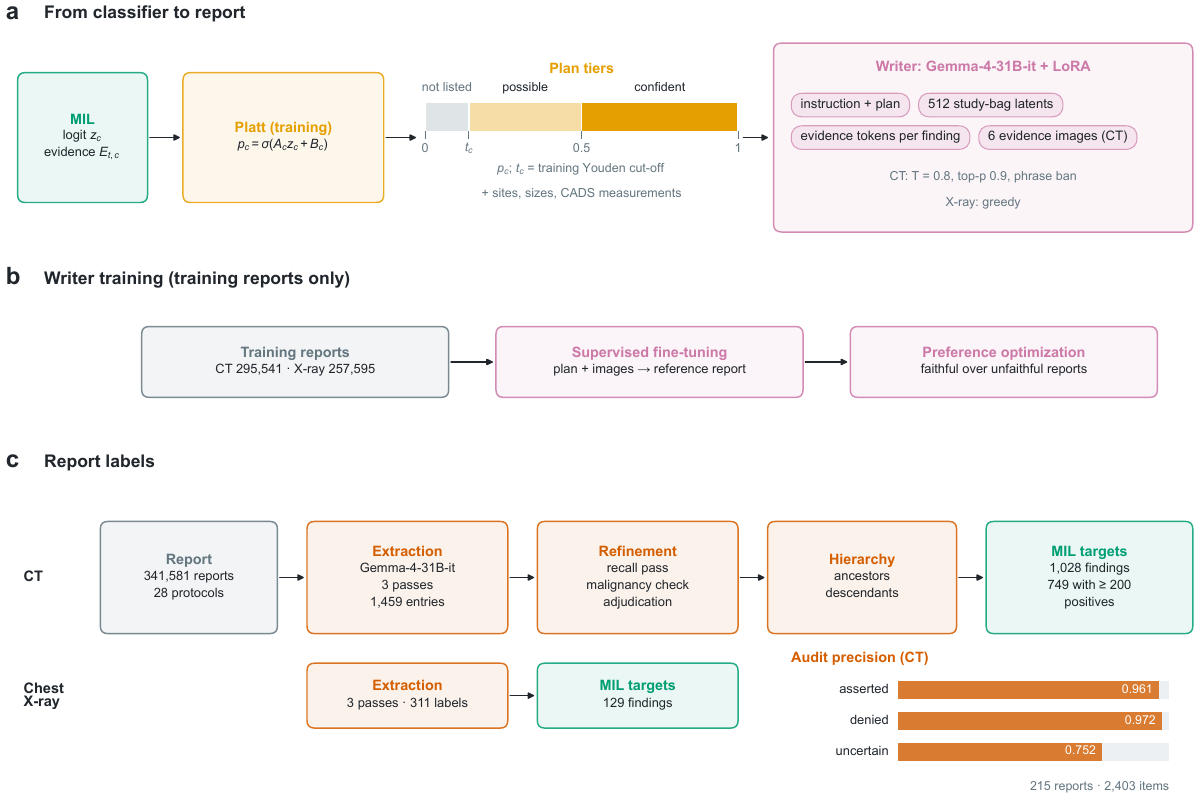}\par}
  \caption[Report generation and report labels.]{Report generation and report labels. \textbf{a}, From classifier probabilities to the report plan and the writer. \textbf{b}, Writer training. \textbf{c}, Report-label extraction and its audit.}
  \label{fig:supp-reportgen-labels}
\end{supplementaryfigure}
\clearpage
\endgroup
\restoregeometry
\pagestyle{forithmus}

%% file: s3-prompts-ct-training-labels-float.tex
\clearpage
\newgeometry{left=25mm,right=25mm,top=17.3mm,bottom=19.0mm,headsep=3.5mm,footskip=4.5mm}
\begingroup
\pagestyle{forithmus}
\setlength{\parskip}{0pt}
\makeatletter
\setlength{\@fptop}{0pt}
\setlength{\@fpbot}{0pt plus 1fil}
\makeatother
\begin{supplementaryfigure}[p]
  {\centering\includegraphics[width=\textwidth,height=\dimexpr\textheight-21.6mm\relax,keepaspectratio,page=1]{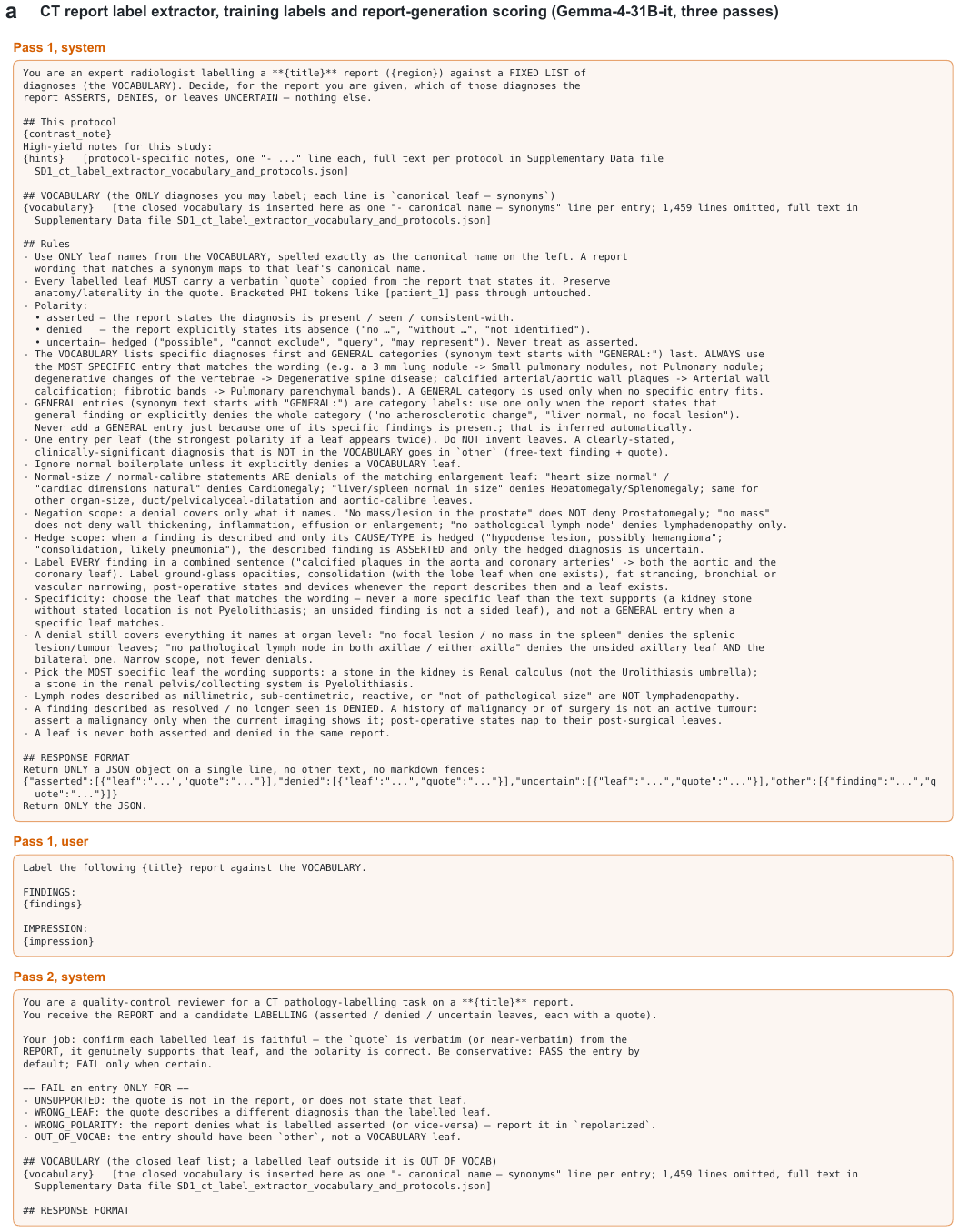}\par}
  \caption[Prompts of the report label extractors.]{Prompts of the report label extractors. \textbf{a}, Three-pass CT label extractor, used for the CT training labels and to score CT report generation. \textbf{b}, Refinement passes of the CT training labels. \textbf{c}, Chest X-ray label extractor, used for the chest radiograph training labels and to score chest radiograph report generation.}
  \label{fig:supp-prompts-labels}
\end{supplementaryfigure}
\clearpage
\begin{supplementaryfigure}[p]
  \centering
  \includegraphics[width=\textwidth,height=\dimexpr\textheight-8mm\relax,keepaspectratio,page=2]{s3-prompts-ct-training-labels.pdf}\par\vspace{1mm}
  {\small\textsf{Supplementary Figure~\ref{fig:supp-prompts-labels}, page 2 of 6}}
\end{supplementaryfigure}
\clearpage
\begin{supplementaryfigure}[p]
  \centering
  \includegraphics[width=\textwidth,height=\dimexpr\textheight-8mm\relax,keepaspectratio,page=3]{s3-prompts-ct-training-labels.pdf}\par\vspace{1mm}
  {\small\textsf{Supplementary Figure~\ref{fig:supp-prompts-labels}, page 3 of 6}}
\end{supplementaryfigure}
\clearpage
\begin{supplementaryfigure}[p]
  \centering
  \includegraphics[width=\textwidth,height=\dimexpr\textheight-8mm\relax,keepaspectratio,page=4]{s3-prompts-ct-training-labels.pdf}\par\vspace{1mm}
  {\small\textsf{Supplementary Figure~\ref{fig:supp-prompts-labels}, page 4 of 6}}
\end{supplementaryfigure}
\clearpage
\begin{supplementaryfigure}[p]
  \centering
  \includegraphics[width=\textwidth,height=\dimexpr\textheight-8mm\relax,keepaspectratio,page=5]{s3-prompts-ct-training-labels.pdf}\par\vspace{1mm}
  {\small\textsf{Supplementary Figure~\ref{fig:supp-prompts-labels}, page 5 of 6}}
\end{supplementaryfigure}
\clearpage
\begin{supplementaryfigure}[p]
  \centering
  \includegraphics[width=\textwidth,height=\dimexpr\textheight-8mm\relax,keepaspectratio,page=6]{s3-prompts-ct-training-labels.pdf}\par\vspace{1mm}
  {\small\textsf{Supplementary Figure~\ref{fig:supp-prompts-labels}, page 6 of 6}}
\end{supplementaryfigure}
\clearpage
\endgroup
\restoregeometry
\pagestyle{forithmus}

%% file: s4-architecture-float.tex
\clearpage
\newgeometry{left=25mm,right=25mm,top=17.3mm,bottom=19.0mm,headsep=3.5mm,footskip=4.5mm}
\begingroup
\pagestyle{forithmus}
\setlength{\parskip}{0pt}
\makeatletter
\setlength{\@fptop}{0pt}
\setlength{\@fpbot}{0pt plus 1fil}
\makeatother
\begin{supplementaryfigure}[p]
  {\centering\includegraphics[width=\textwidth,height=\dimexpr\textheight-13.8mm\relax,keepaspectratio]{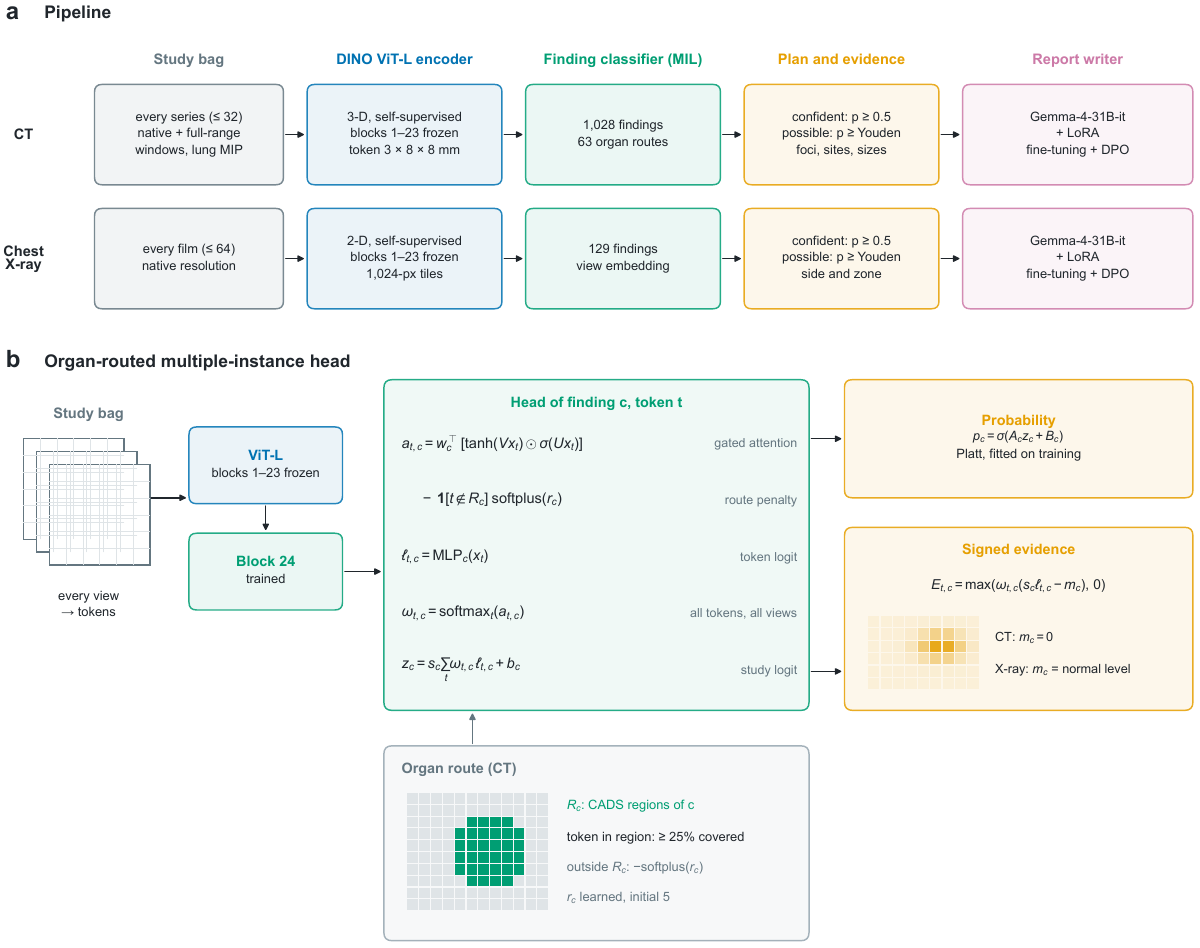}\par}
  \caption[Architecture of FORA.]{Architecture of FORA. \textbf{a}, Pipeline for CT and chest X-ray. \textbf{b}, Organ-routed multiple-instance head with attention, study logit and signed evidence.}
  \label{fig:supp-architecture}
\end{supplementaryfigure}
\clearpage
\endgroup
\restoregeometry
\pagestyle{forithmus}

%% file: s5-evaluation-protocol-float.tex
\clearpage
\newgeometry{left=25mm,right=25mm,top=17.3mm,bottom=19.0mm,headsep=3.5mm,footskip=4.5mm}
\begingroup
\pagestyle{forithmus}
\setlength{\parskip}{0pt}
\makeatletter
\setlength{\@fptop}{0pt}
\setlength{\@fpbot}{0pt plus 1fil}
\makeatother
\begin{supplementaryfigure}[p]
  {\centering\includegraphics[width=\textwidth,height=\dimexpr\textheight-13.8mm\relax,keepaspectratio]{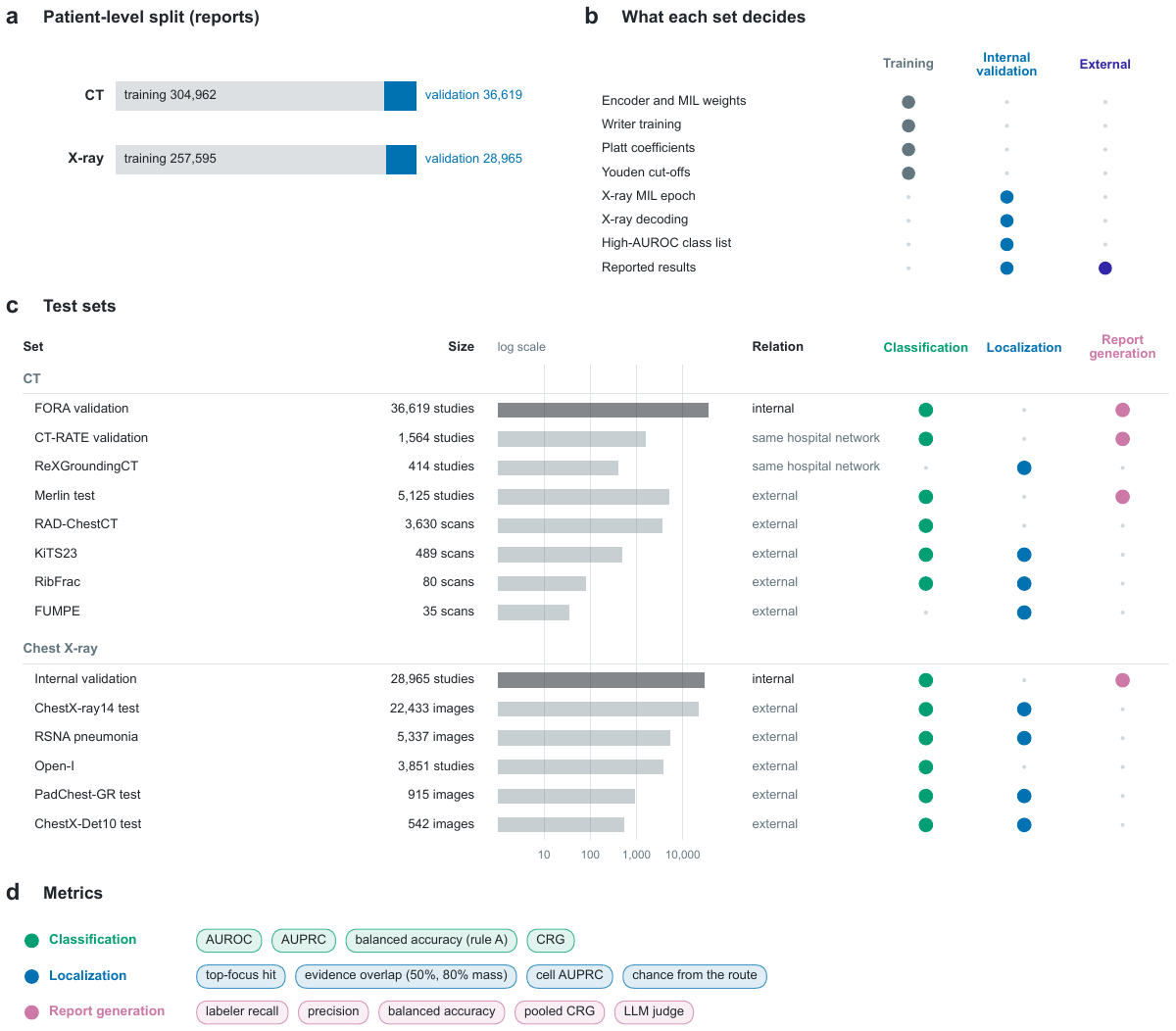}\par}
  \caption[Evaluation protocol.]{Evaluation protocol. \textbf{a}, Patient-level split. \textbf{b}, What each set is used for. \textbf{c}, Test sets with their size, relation to the training data and tasks. \textbf{d}, Metrics for each task.}
  \label{fig:supp-evaluation}
\end{supplementaryfigure}
\clearpage
\endgroup
\restoregeometry
\pagestyle{forithmus}

%% file: s6-classifier-breadth-training-data-float.tex
\clearpage
\newgeometry{left=25mm,right=25mm,top=17.3mm,bottom=19.0mm,headsep=3.5mm,footskip=4.5mm}
\begingroup
\pagestyle{forithmus}
\setlength{\parskip}{0pt}
\makeatletter
\setlength{\@fptop}{0pt}
\setlength{\@fpbot}{0pt plus 1fil}
\makeatother
\begin{supplementaryfigure}[p]
  {\centering\includegraphics[width=\textwidth,height=\dimexpr\textheight-21.6mm\relax,keepaspectratio]{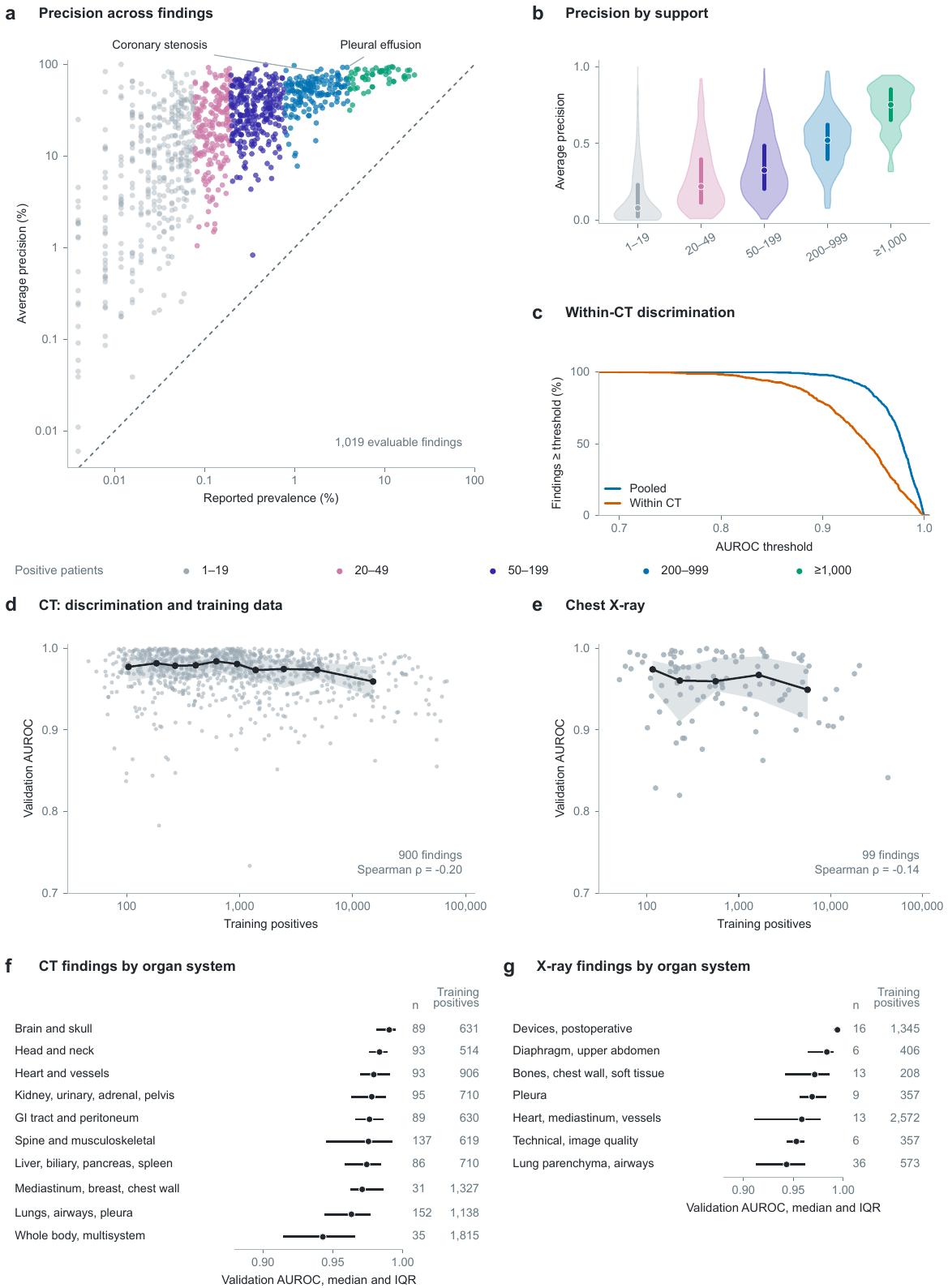}\par}
  \caption[Classifier breadth and training data.]{Classifier breadth and training data. \textbf{a}, Average precision against prevalence for the 1,019 CT findings with at least one positive held-out patient, using one examination per patient. \textbf{b}, Average precision by number of positives. \textbf{c}, Pooled and within-CT-type AUROC. \textbf{d,e}, AUROC against number of training positives for CT (d) and chest X-ray (e). \textbf{f,g}, AUROC by organ system.}
  \label{fig:fora-supp2-classifier-reliability}
\end{supplementaryfigure}
\clearpage
\endgroup
\restoregeometry
\pagestyle{forithmus}

%% file: s7-calibration-float.tex
\clearpage
\newgeometry{left=25mm,right=25mm,top=17.3mm,bottom=19.0mm,headsep=3.5mm,footskip=4.5mm}
\begingroup
\pagestyle{forithmus}
\setlength{\parskip}{0pt}
\makeatletter
\setlength{\@fptop}{0pt}
\setlength{\@fpbot}{0pt plus 1fil}
\makeatother
\begin{supplementaryfigure}[p]
  {\centering\includegraphics[width=\textwidth,height=\dimexpr\textheight-17.7mm\relax,keepaspectratio]{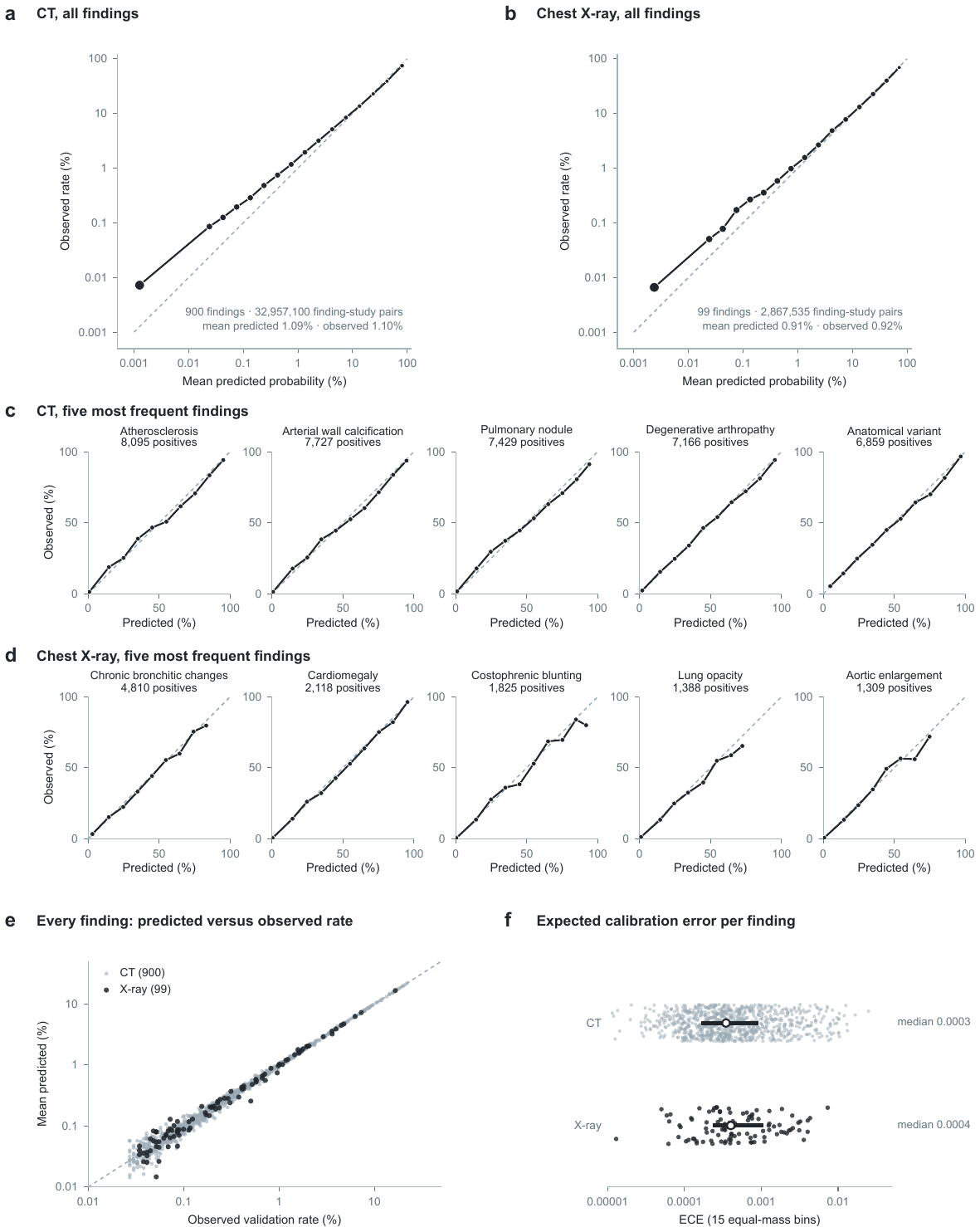}\par}
  \caption[Calibration of the FORA classifiers.]{Calibration of the FORA classifiers. \textbf{a,b}, Reliability over all findings for CT (a) and chest X-ray (b). \textbf{c,d}, The five most frequent findings for CT (c) and chest X-ray (d). \textbf{e}, Predicted against observed rate for each finding. \textbf{f}, Expected calibration error for each finding.}
  \label{fig:supp-calibration}
\end{supplementaryfigure}
\clearpage
\endgroup
\restoregeometry
\pagestyle{forithmus}

%% file: s8-external-transfer-float.tex
\clearpage
\newgeometry{left=25mm,right=25mm,top=17.3mm,bottom=19.0mm,headsep=3.5mm,footskip=4.5mm}
\begingroup
\pagestyle{forithmus}
\setlength{\parskip}{0pt}
\makeatletter
\setlength{\@fptop}{0pt}
\setlength{\@fpbot}{0pt plus 1fil}
\makeatother
\begin{supplementaryfigure}[p]
  {\centering\includegraphics[width=\textwidth,height=\dimexpr\textheight-29.4mm\relax,keepaspectratio]{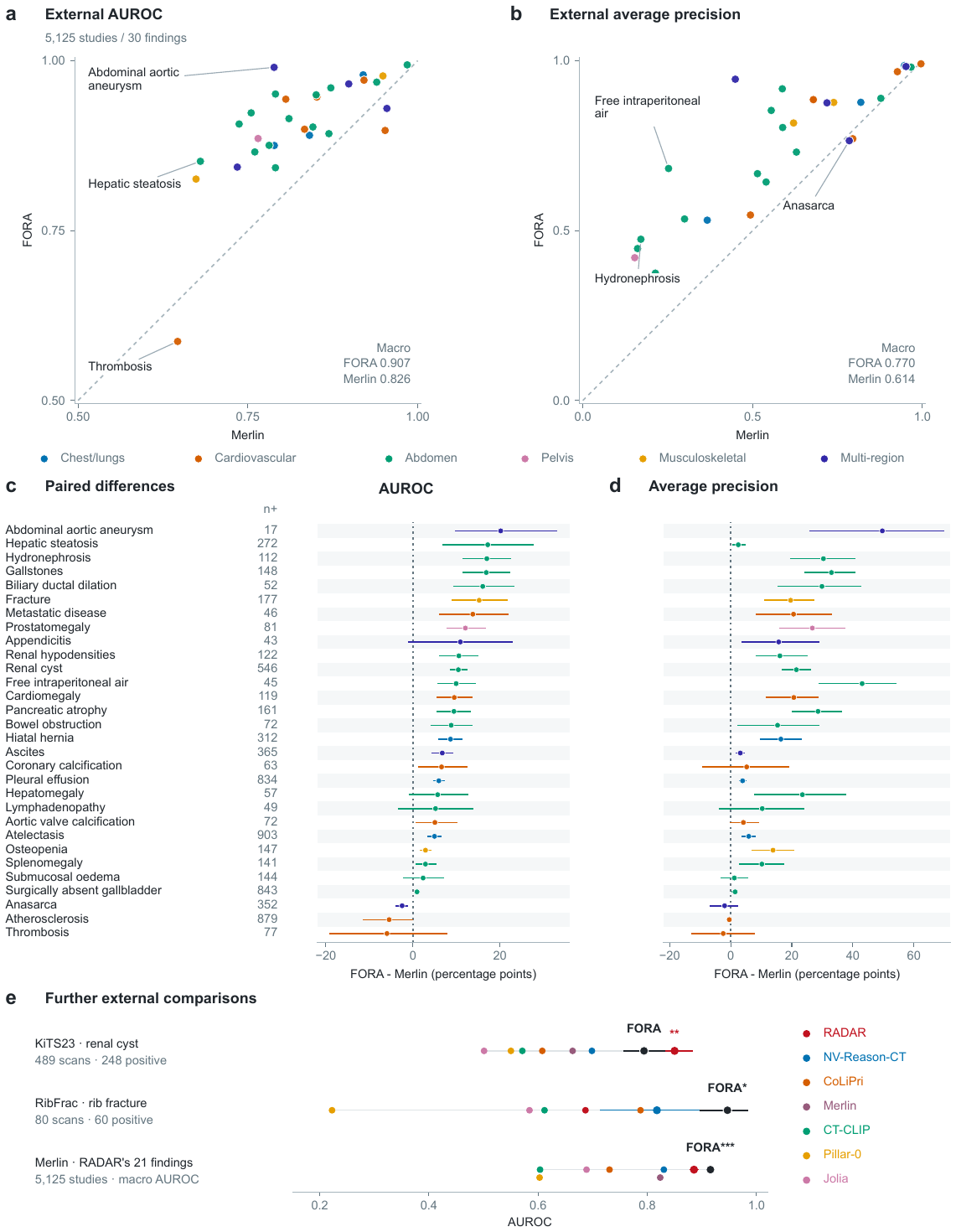}\par}
  \caption[External transfer of the FORA classifier.]{External transfer of the FORA classifier. \textbf{a,b}, AUROC (a) and average precision (b) of FORA and Merlin for 30 findings of the Merlin test set. \textbf{c,d}, FORA minus Merlin for each finding. \textbf{e}, AUROC for renal cyst in KiTS23 and rib fracture in RibFrac, and macro AUROC on the 21 Merlin findings covered by RADAR's released protocol. Error bars, 95\% bootstrap confidence intervals. Stars, two-sided paired tests of FORA against the best other model with Holm correction, shown on the higher of the two ($^{*}P<0.05$, $^{**}P<0.01$, $^{***}P<0.001$).}
  \label{fig:fora-supp3-transfer-training}
\end{supplementaryfigure}
\clearpage
\endgroup
\restoregeometry
\pagestyle{forithmus}

%% file: s9-subgroup-robustness-float.tex
\clearpage
\newgeometry{left=25mm,right=25mm,top=17.3mm,bottom=19.0mm,headsep=3.5mm,footskip=4.5mm}
\begingroup
\pagestyle{forithmus}
\setlength{\parskip}{0pt}
\makeatletter
\setlength{\@fptop}{0pt}
\setlength{\@fpbot}{0pt plus 1fil}
\makeatother
\begin{supplementaryfigure}[p]
  {\centering\includegraphics[width=\textwidth,height=\dimexpr\textheight-17.7mm\relax,keepaspectratio]{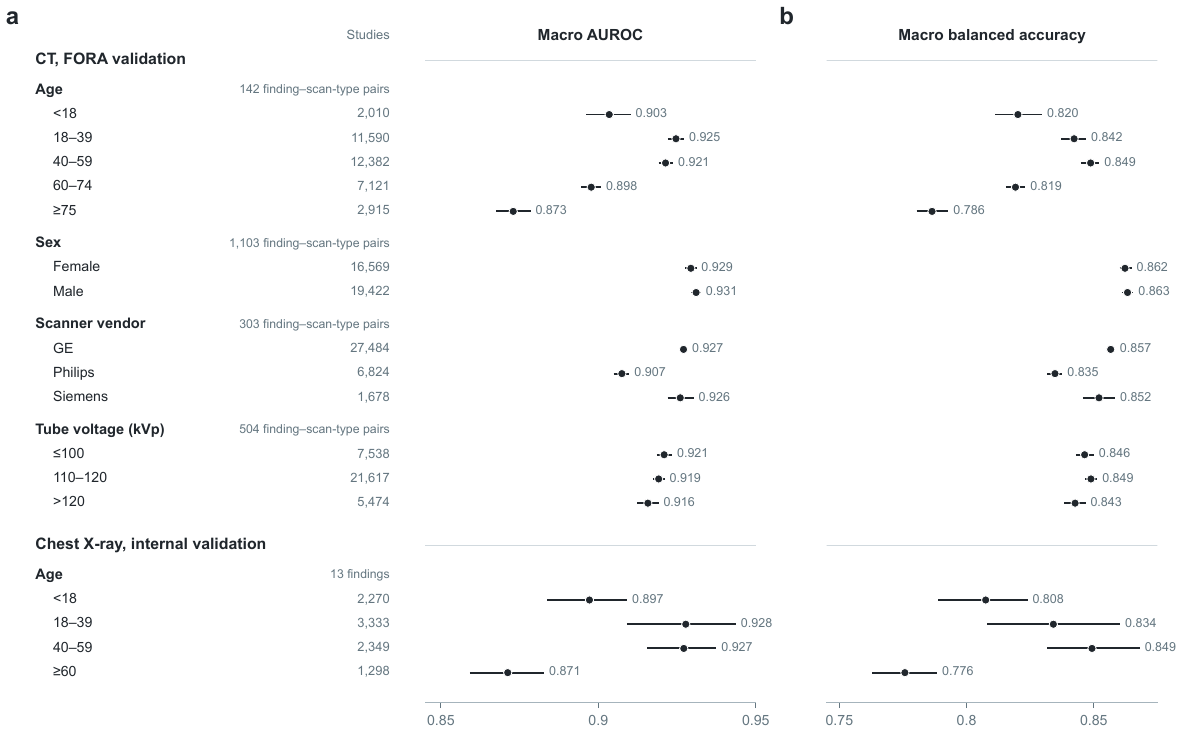}\par}
  \caption[Subgroup robustness of the FORA classifier.]{Subgroup robustness of the FORA classifier. \textbf{a,b}, Macro AUROC (a) and balanced accuracy (b) by age, sex, scanner vendor and tube voltage for CT, and by age for chest X-ray. Error bars, 95\% bootstrap confidence intervals.}
  \label{fig:supp-subgroups}
\end{supplementaryfigure}
\clearpage
\endgroup
\restoregeometry
\pagestyle{forithmus}

%% file: s10-grounded-findings-gallery-float.tex
\clearpage
\newgeometry{left=25mm,right=25mm,top=17.3mm,bottom=19.0mm,headsep=3.5mm,footskip=4.5mm}
\begingroup
\pagestyle{forithmus}
\setlength{\parskip}{0pt}
\makeatletter
\setlength{\@fptop}{0pt}
\setlength{\@fpbot}{0pt plus 1fil}
\makeatother
\begin{supplementaryfigure}[p]
  {\centering\includegraphics[width=\textwidth,height=\dimexpr\textheight-17.7mm\relax,keepaspectratio]{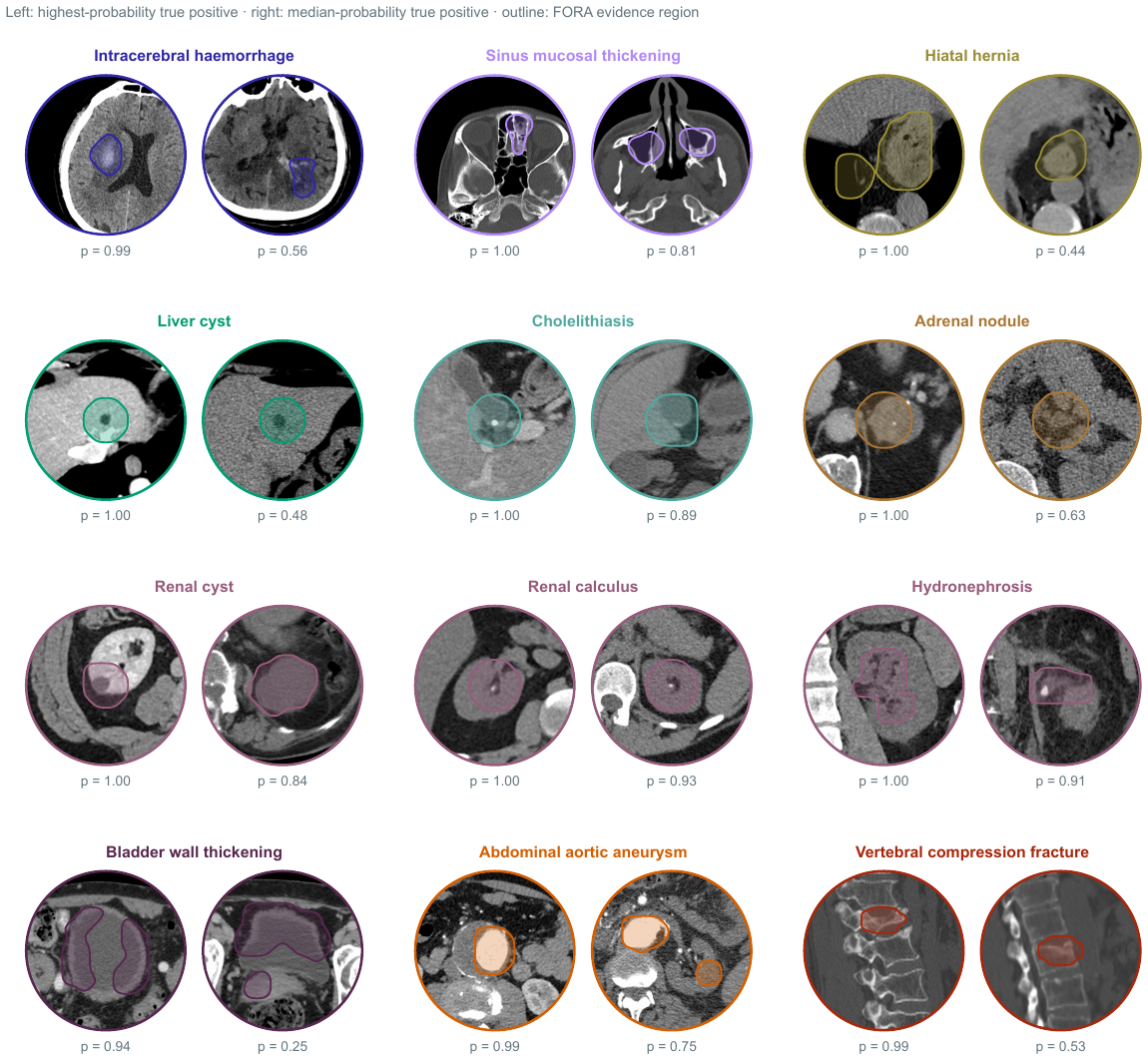}\par}
  \caption[Grounded findings outside the chest in FORA validation studies.]{Grounded findings outside the chest in FORA validation studies. For each finding, the highest-probability (left) and median-probability (right) true positive, with FORA's evidence region outlined, and p is the FORA probability.}
  \label{fig:supp-gallery}
\end{supplementaryfigure}
\clearpage
\endgroup
\restoregeometry
\pagestyle{forithmus}

%% file: s11-prompts-report-writers-float.tex
\clearpage
\newgeometry{left=25mm,right=25mm,top=17.3mm,bottom=19.0mm,headsep=3.5mm,footskip=4.5mm}
\begingroup
\pagestyle{forithmus}
\setlength{\parskip}{0pt}
\makeatletter
\setlength{\@fptop}{0pt}
\setlength{\@fpbot}{0pt plus 1fil}
\makeatother
\begin{supplementaryfigure}[p]
  {\centering\includegraphics[width=\textwidth,height=\dimexpr\textheight-9.9mm\relax,keepaspectratio]{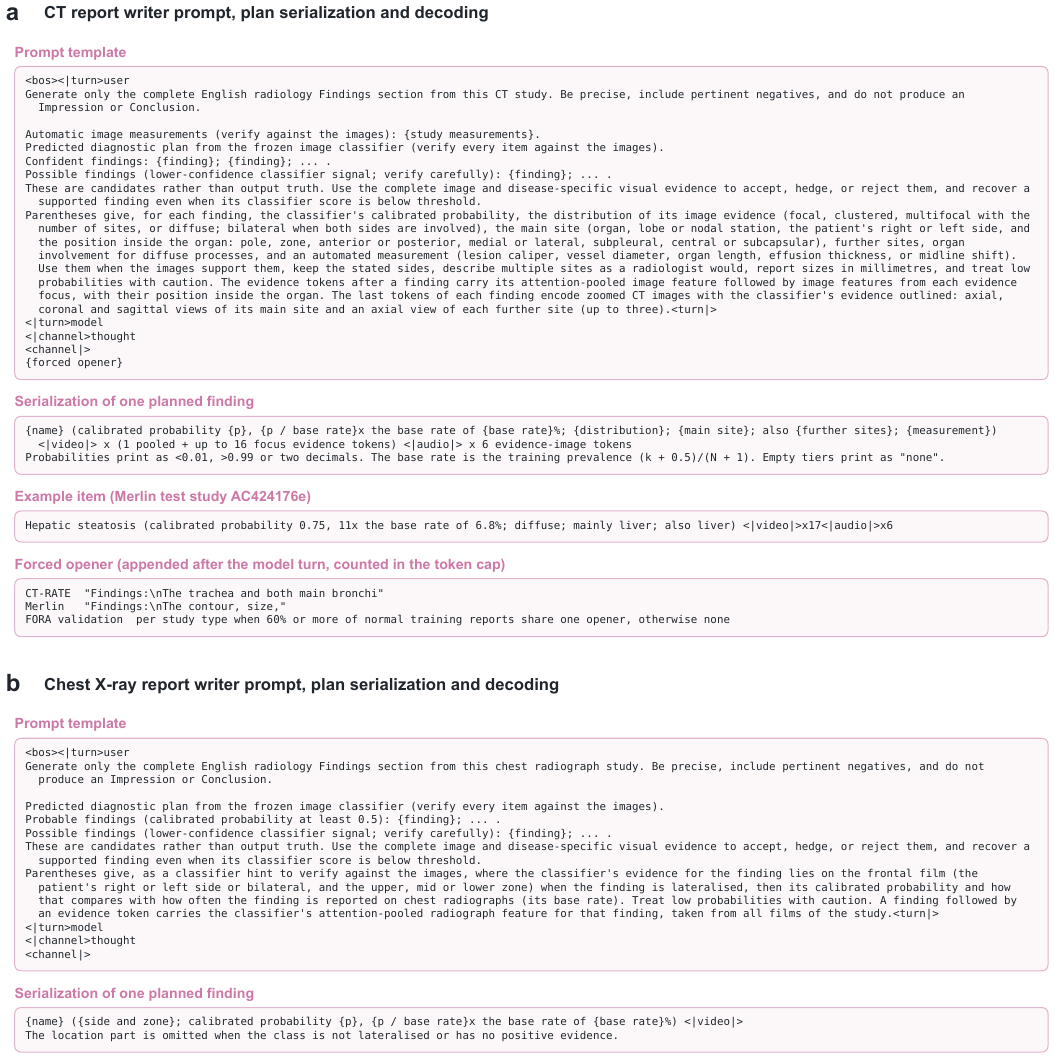}\par}
  \caption[Prompts of the report writers.]{Prompts of the report writers. \textbf{a}, CT. \textbf{b}, Chest X-ray.}
  \label{fig:supp-prompts-writers}
\end{supplementaryfigure}
\clearpage
\endgroup
\restoregeometry
\pagestyle{forithmus}

%% file: s12-dpo-ablation-float.tex
\clearpage
\newgeometry{left=25mm,right=25mm,top=17.3mm,bottom=19.0mm,headsep=3.5mm,footskip=4.5mm}
\begingroup
\pagestyle{forithmus}
\setlength{\parskip}{0pt}
\makeatletter
\setlength{\@fptop}{0pt}
\setlength{\@fpbot}{0pt plus 1fil}
\makeatother
\begin{supplementaryfigure}[p]
  {\centering\includegraphics[width=\textwidth,height=\dimexpr\textheight-37.2mm\relax,keepaspectratio]{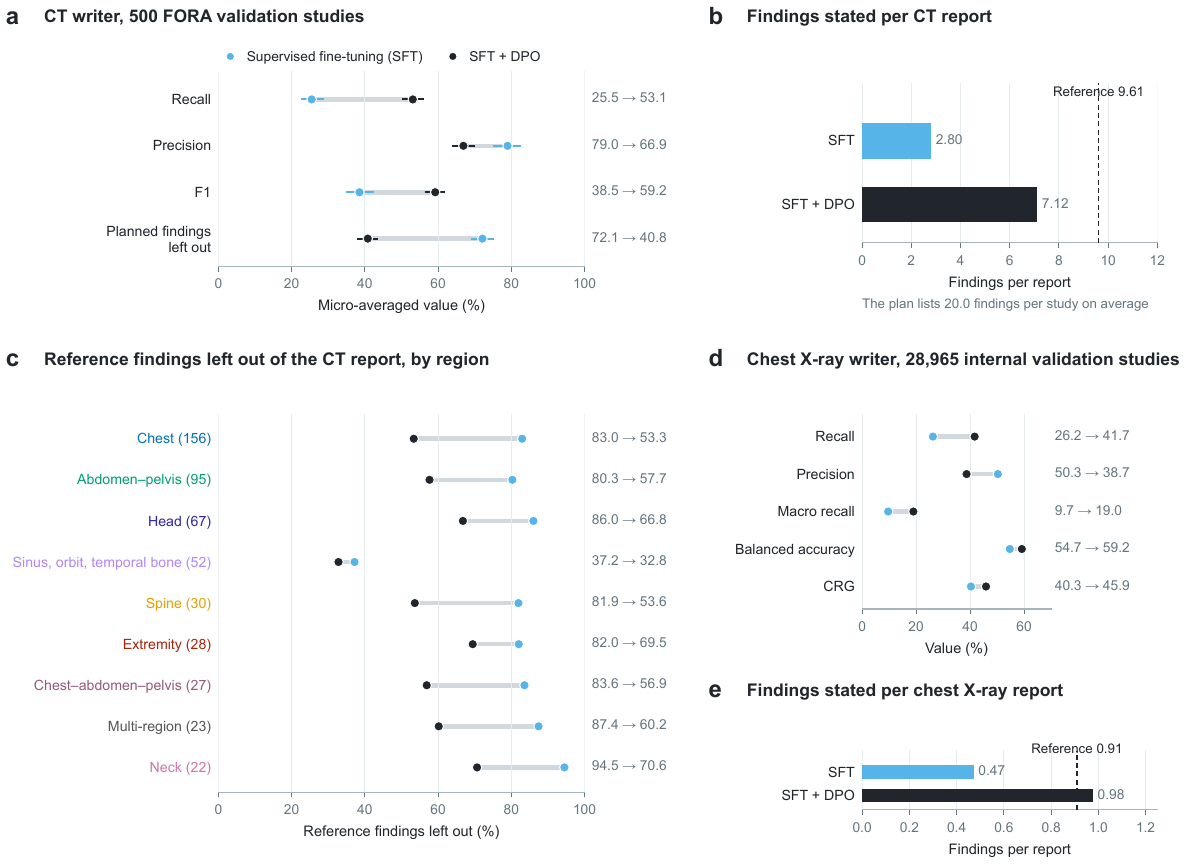}\par}
  \caption[Preference optimization of the report writers.]{Preference optimization of the report writers. \textbf{a}, Micro-averaged recall, precision and F1 of the CT writer after supervised fine-tuning (SFT, blue) and after direct preference optimization (DPO) from the same checkpoint (black), and the share of reference findings in the plan that the report leaves out, on 500 FORA validation studies. \textbf{b}, Findings stated per CT report. Dashed line, reference reports. \textbf{c}, Reference findings left out of the CT report by region, over all 1,028 findings (number of studies in brackets). \textbf{d}, The chest X-ray writer after SFT and after DPO on all 28,965 internal validation studies. \textbf{e}, Findings stated per chest X-ray report. Plans, references and scoring are identical before and after DPO. Error bars, 95\% bootstrap confidence intervals.}
  \label{fig:supp-dpo}
\end{supplementaryfigure}
\clearpage
\endgroup
\restoregeometry
\pagestyle{forithmus}

%% file: s14-prompts-judge-baselines-float.tex
\clearpage
\newgeometry{left=25mm,right=25mm,top=17.3mm,bottom=19.0mm,headsep=3.5mm,footskip=4.5mm}
\begingroup
\pagestyle{forithmus}
\setlength{\parskip}{0pt}
\makeatletter
\setlength{\@fptop}{0pt}
\setlength{\@fpbot}{0pt plus 1fil}
\makeatother
\begin{supplementaryfigure}[p]
  {\centering\includegraphics[width=\textwidth,height=\dimexpr\textheight-13.8mm\relax,keepaspectratio]{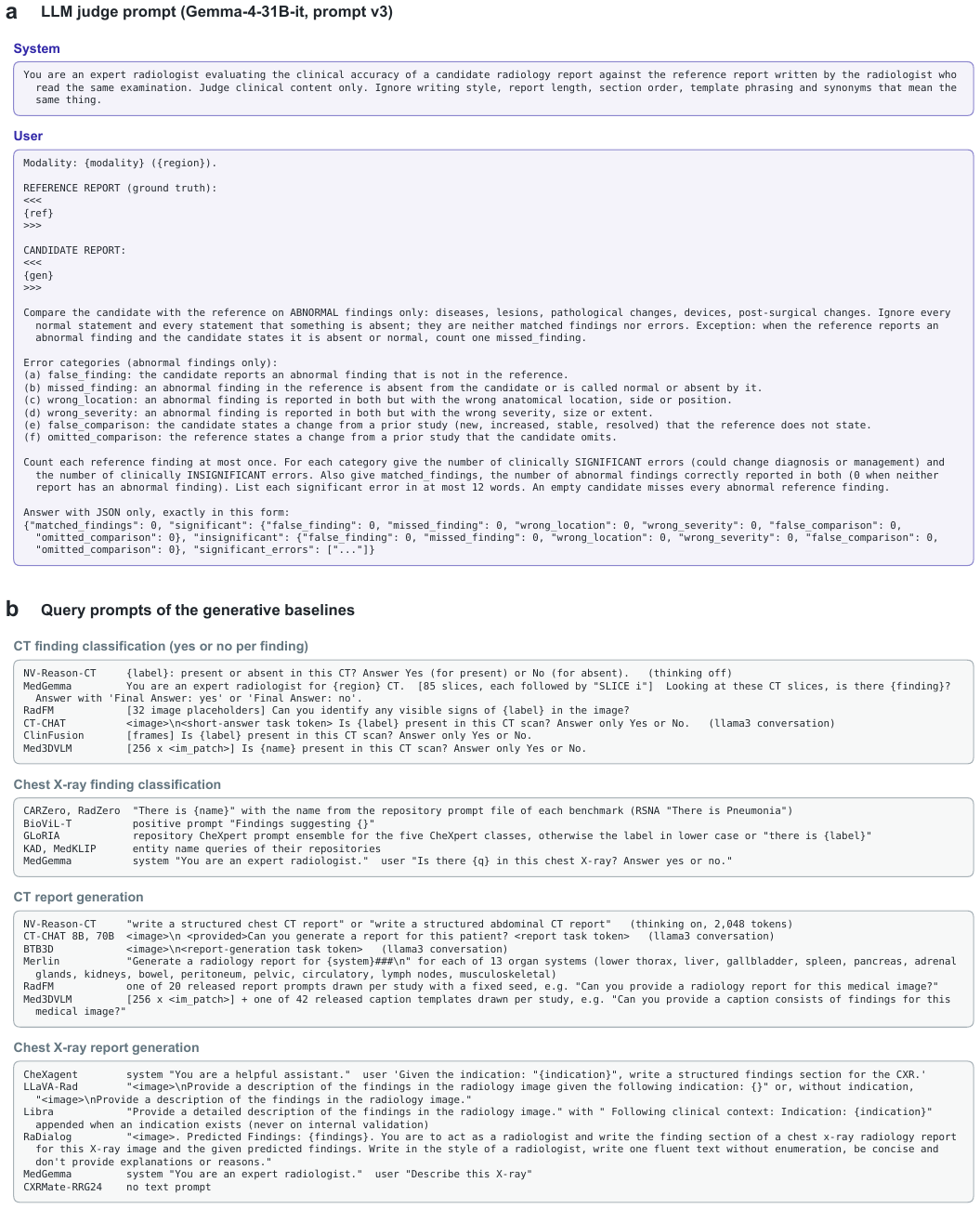}\par}
  \caption[Prompts of the LLM judge and the other report generators.]{Prompts of the LLM judge and the other report generators. \textbf{a}, LLM judge. \textbf{b}, Other report generators.}
  \label{fig:supp-prompts-judge}
\end{supplementaryfigure}
\clearpage
\endgroup
\restoregeometry
\pagestyle{forithmus}

%% file: s15-report-examples-float.tex
\clearpage
\newgeometry{left=25mm,right=25mm,top=17.3mm,bottom=19.0mm,headsep=3.5mm,footskip=4.5mm}
\begingroup
\pagestyle{forithmus}
\setlength{\parskip}{0pt}
\makeatletter
\setlength{\@fptop}{0pt}
\setlength{\@fpbot}{0pt plus 1fil}
\makeatother
\begin{supplementaryfigure}[p]
  {\centering\includegraphics[width=\textwidth,height=\dimexpr\textheight-21.6mm\relax,keepaspectratio,page=1]{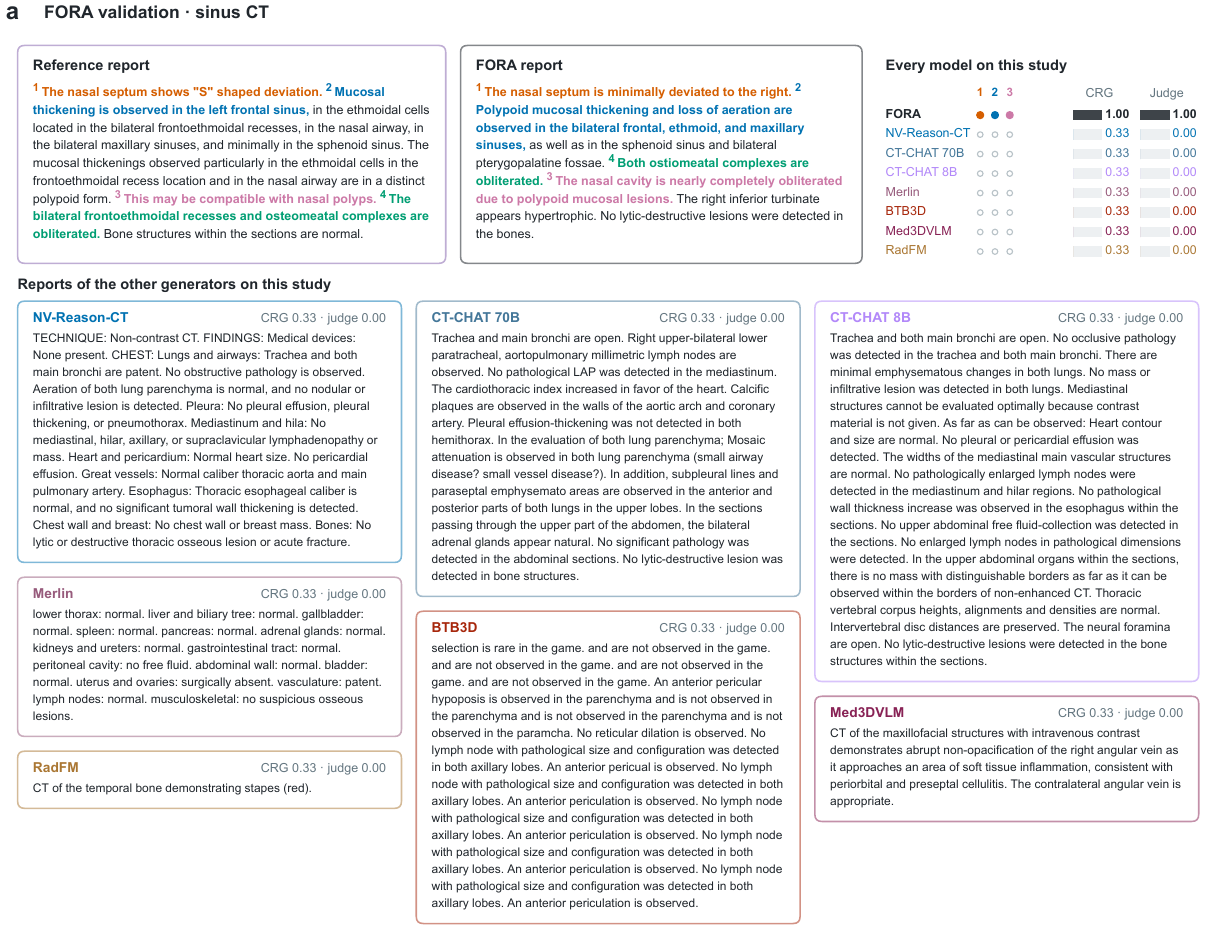}\par}
  \caption[Report generation examples.]{Report generation examples. Each page shows one study with the reference report, FORA's report with matching statements numbered, the report of each other generator, and their CRG and judge scores. \textbf{a}, FORA validation sinus CT. \textbf{b}, FORA validation abdominal CT. \textbf{c}, CT-RATE chest CT. \textbf{d}, Merlin abdominal CT.}
  \label{fig:supp-report-examples}
\end{supplementaryfigure}
\clearpage
\begin{supplementaryfigure}[p]
  \centering
  \includegraphics[width=\textwidth,height=\dimexpr\textheight-8mm\relax,keepaspectratio,page=2]{s15-report-examples.pdf}\par\vspace{1mm}
  {\small\textsf{Supplementary Figure~\ref{fig:supp-report-examples}, page 2 of 4}}
\end{supplementaryfigure}
\clearpage
\begin{supplementaryfigure}[p]
  \centering
  \includegraphics[width=\textwidth,height=\dimexpr\textheight-8mm\relax,keepaspectratio,page=3]{s15-report-examples.pdf}\par\vspace{1mm}
  {\small\textsf{Supplementary Figure~\ref{fig:supp-report-examples}, page 3 of 4}}
\end{supplementaryfigure}
\clearpage
\begin{supplementaryfigure}[p]
  \centering
  \includegraphics[width=\textwidth,height=\dimexpr\textheight-8mm\relax,keepaspectratio,page=4]{s15-report-examples.pdf}\par\vspace{1mm}
  {\small\textsf{Supplementary Figure~\ref{fig:supp-report-examples}, page 4 of 4}}
\end{supplementaryfigure}
\clearpage
\endgroup
\restoregeometry
\pagestyle{forithmus}

%% file: s16-quantitative-tissue-float.tex
\clearpage
\newgeometry{left=25mm,right=25mm,top=17.3mm,bottom=19.0mm,headsep=3.5mm,footskip=4.5mm}
\begingroup
\pagestyle{forithmus}
\setlength{\parskip}{0pt}
\makeatletter
\setlength{\@fptop}{0pt}
\setlength{\@fpbot}{0pt plus 1fil}
\makeatother
\begin{supplementaryfigure}[p]
  {\centering\includegraphics[width=\textwidth,height=\dimexpr\textheight-25.5mm\relax,keepaspectratio]{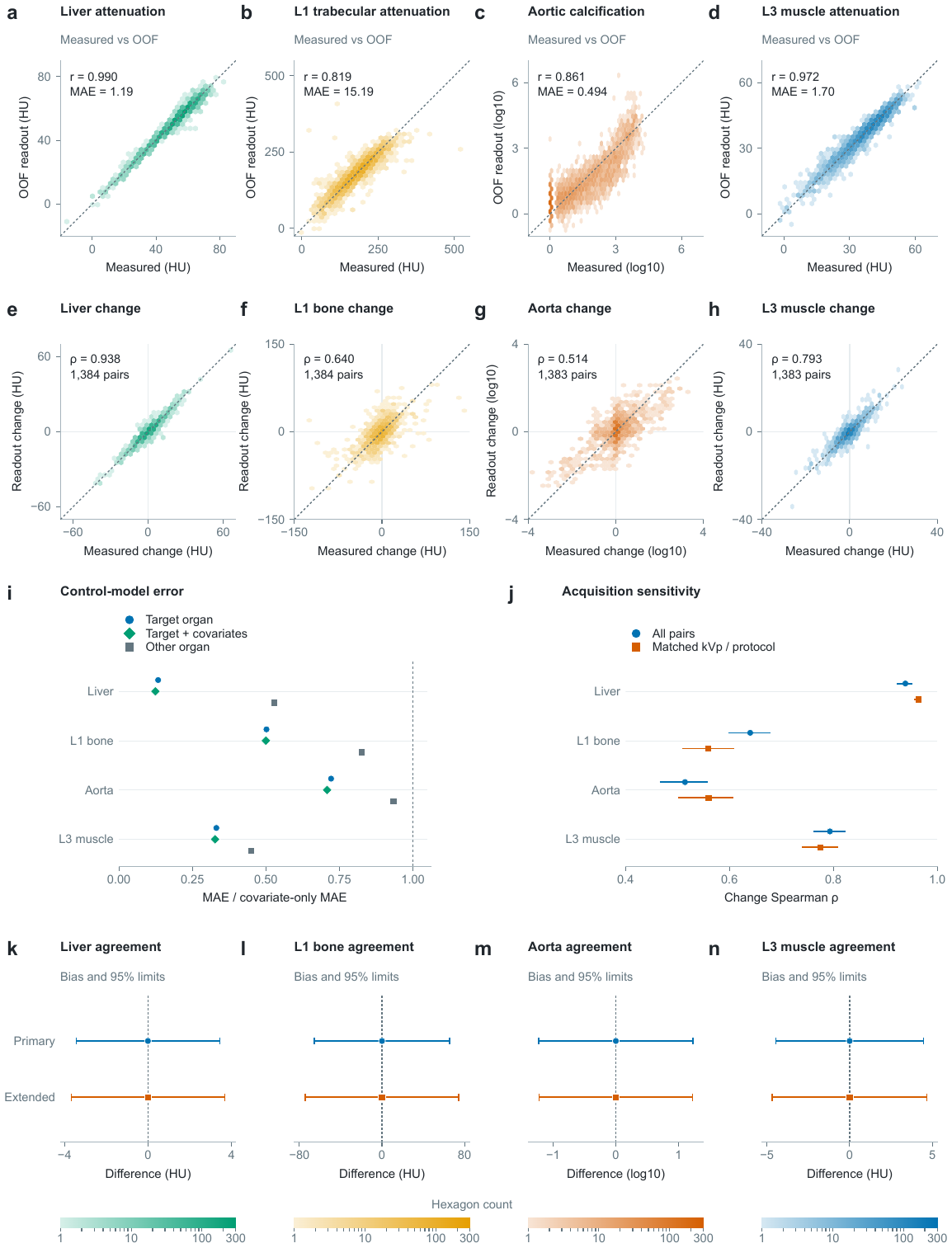}\par}
  \caption[Continuous tissue information in FORA features.]{Continuous tissue information in FORA features. \textbf{a--d}, Measured liver attenuation, L1 bone attenuation, aortic calcification and L3 muscle attenuation against out-of-fold readouts. \textbf{e--h}, Within-patient changes between two scans. \textbf{i}, Error of target-organ, other-organ and demographic readouts. \textbf{j}, Correlation of measured and readout changes. \textbf{k--n}, Bland--Altman agreement. Error bars, 95\% patient-bootstrap confidence intervals.}
  \label{fig:fora-supp4-quantitative-information}
\end{supplementaryfigure}
\clearpage
\endgroup
\restoregeometry
\pagestyle{forithmus}

%% file: s17-repeat-consistency-float.tex
\clearpage
\newgeometry{left=25mm,right=25mm,top=17.3mm,bottom=19.0mm,headsep=3.5mm,footskip=4.5mm}
\begingroup
\pagestyle{forithmus}
\setlength{\parskip}{0pt}
\makeatletter
\setlength{\@fptop}{0pt}
\setlength{\@fpbot}{0pt plus 1fil}
\makeatother
\begin{supplementaryfigure}[p]
  {\centering\includegraphics[width=\textwidth,height=\dimexpr\textheight-25.5mm\relax,keepaspectratio]{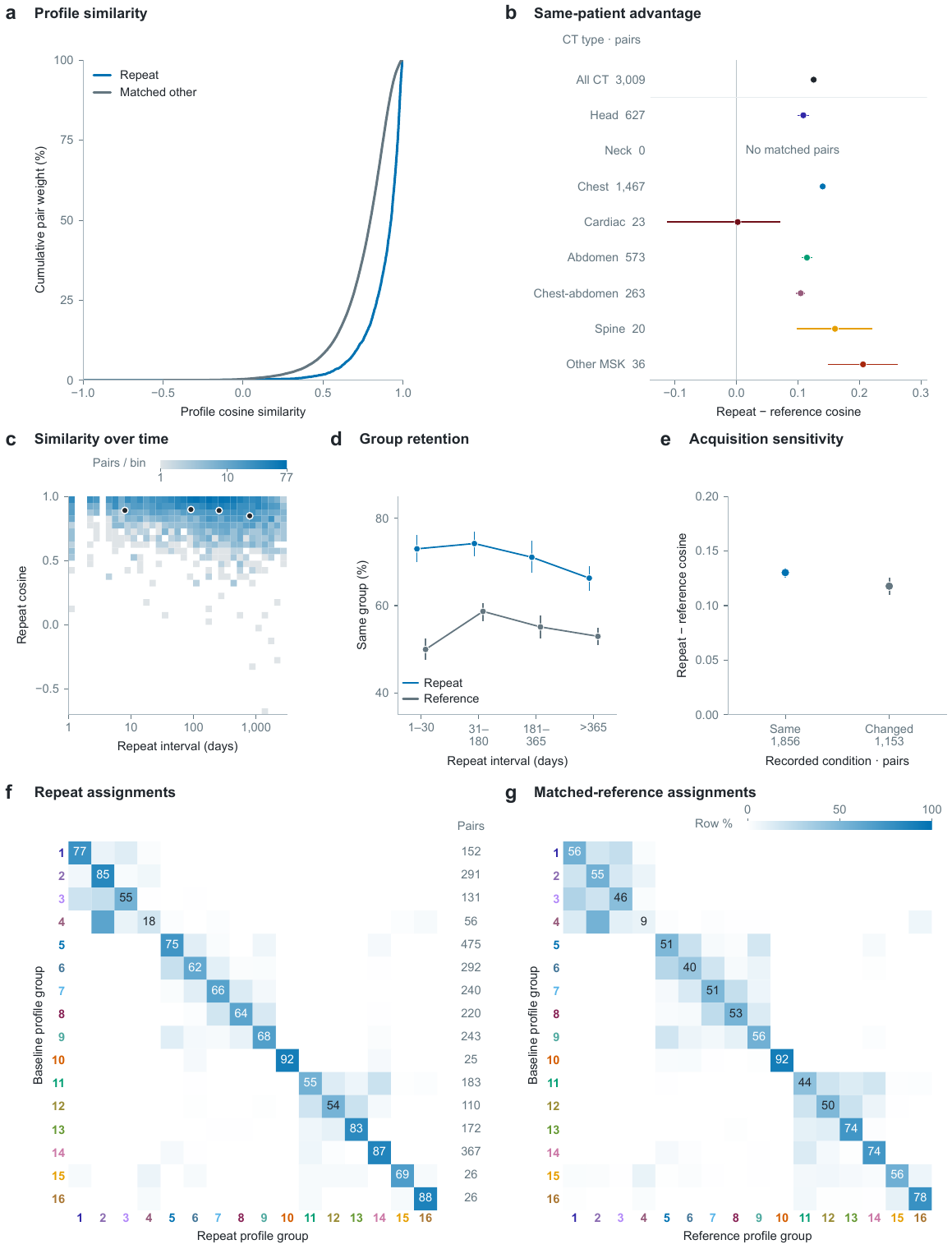}\par}
  \caption[Consistency of CT profiles across repeat examinations.]{Consistency of CT profiles across repeat examinations. \textbf{a}, Similarity of each examination to the same patient's repeat (blue) and to matched other patients (grey). \textbf{b}, Repeat advantage by CT category. \textbf{c}, Repeat similarity by interval. \textbf{d}, Group retention by interval. \textbf{e}, Repeat advantage for same and changed site, contrast or manufacturer. \textbf{f,g}, Group assignment of repeat (f) and reference (g) examinations, groups as in Extended Data Fig.~\ref{fig:fora-ed2-global-atlas}. Error bars, 95\% bootstrap confidence intervals.}
  \label{fig:fora-supp5-repeat-consistency}
\end{supplementaryfigure}
\clearpage
\endgroup
\restoregeometry
\pagestyle{forithmus}